\documentclass[preprint,12pt]{elsarticle}

\usepackage{amssymb}
\usepackage{amsmath}
\usepackage{amsfonts}
\usepackage{algorithmic}
\usepackage{algorithm}
\usepackage{array}
\usepackage{amsthm}
\usepackage[caption=false,font=normalsize,labelfont=sf,textfont=sf]{subfig}
\usepackage{textcomp}
\usepackage{stfloats}
\usepackage{url}
\usepackage{adjustbox}
\usepackage{verbatim}
\usepackage{multirow} 
\usepackage{graphicx}
\usepackage{multirow}
\usepackage[table,xcdraw]{xcolor}
\usepackage{colortbl}
\usepackage[normalem]{ulem}
\usepackage{bm}
\useunder{\uline}{\ul}{}
\newtheorem{remark}{\bf Remark}

\journal{}

\begin{document}

\begin{frontmatter}



\title{Communication-Efficient Personalized Federal Graph Learning via Low-Rank Decomposition}


\author[label1,label2]{Ruyue Liu} 
\ead{liuruyue@iie.ac.cn}
\author[label1,label2]{Rong Yin\corref{cor1}}
\ead{yinrong@iie.ac.cn}
\author[label3]{Xiangzhen Bo}
\ead{353145@whut.edu.cn}
\author[label4]{Xiaoshuai Hao}
\ead{xshao@baai.ac.cn}
\author[label5]{Xingrui Zhou}
\ead{22009100548@stu.xidian.edu.cn}
\author[label6]{Yong Liu}
\ead{liuyonggsai@ruc.edu.cn}
\author[label1]{Can Ma}
\ead{macan@iie.ac.cn}
\author[label1]{Weiping Wang}
\ead{wangweiping@iie.ac.cn}

\cortext[cor1]{Rong Yin is the corresponding author.}

\affiliation[label1]{organization={Institute of Information Engineering, Chinese Academy of Sciences},
            city={Beijing},
            postcode={100085}, 
            country={China}}
\affiliation[label2]{organization={School of Cyberspace Security, University of Chinese Academy of Sciences},
            city={Beijing},
            postcode={100049}, 
            country={China}}
\affiliation[label3]{organization={Wuhan University of Technology},
            city={Wuhan},
            postcode={430070}, 
            country={China}}
\affiliation[label4]{organization={Beijing Academy of Artificial Intelligence},
            city={Beijing},
            postcode={100086}, 
            country={China}}
\affiliation[label5]{organization={Xidian University},
            city={Xi'an},
            postcode={710126}, 
            country={China}}
\affiliation[label6]{organization={Renmin University of China},
            city={Beijing},
            postcode={100872}, 
            country={China}}

\begin{abstract}
Federated graph learning (FGL) has gained significant attention for enabling heterogeneous clients to process their private graph data locally while interacting with a centralized server, thus maintaining privacy. However, graph data on clients are typically non-IID, posing a challenge for a single model to perform well across all clients. Another major bottleneck of FGL is the high cost of communication. To address these challenges, we propose a communication-efficient personalized federated graph learning algorithm, \textbf{CEFGL}. Our method decomposes the model parameters into low-rank generic and sparse private models. We employ a dual-channel encoder to learn sparse local knowledge in a personalized manner and low-rank global knowledge in a shared manner. Additionally, we perform multiple local stochastic gradient descent iterations between communication phases and integrate efficient compression techniques into the algorithm. The advantage of CEFGL lies in its ability to capture common and individual knowledge more precisely. By utilizing low-rank and sparse parameters along with compression techniques, CEFGL significantly reduces communication complexity. Extensive experiments demonstrate that our method achieves optimal classification accuracy in a variety of heterogeneous environments across sixteen datasets. Specifically, compared to the state-of-the-art method FedStar, the proposed method (with GIN as the base model) improves accuracy by 5.64\% on cross-datasets setting CHEM, reduces communication bits by a factor of 18.58, and reduces the communication time by a factor of 1.65.
\end{abstract}

\begin{keyword}
Federated Graph Learning, Low-Rank, Non-IID.
\end{keyword}

\end{frontmatter}



\section{Introduction}
In the modern era of pervasive data generation, graphs have emerged as a powerful tool for representing complex, relational information across various domains, such as social networks, healthcare, finance, and transportation. As data becomes increasingly decentralized, particularly across heterogeneous environments, the need to effectively fuse information from multiple, distributed sources has become paramount. Graph Neural Networks (GNNs) \cite{kipf2016semi,liu2024aswt} have shown great potential in leveraging the inherent topology of graph structures for deep learning tasks. These techniques have found extensive applications in diverse fields, including drug discovery \cite{Su2023,LIU2024102563,WU2023101909}, neuroscience \cite{ding2023lggnet,GORRIZ2023101945}, social networks \cite{liu2024unbiased,LI2024102837}, knowledge graphs \cite{PENG2025102729,ZHANG2024102581}, recommender systems \cite{10114977,PAZRUZA2024102497,HIMEUR20211}, and traffic flow prediction \cite{ZHU2025102703,XU2024102292}. Despite significant advancements, most methods require centralized storage of large graph datasets on a single machine for training. Due to data security and user privacy concerns, data owners (clients) may be reluctant to share their data, leading to data segregation issues \cite{de2023guide}. Furthermore, graph data stored by different clients often exhibit non-identical and independently distributed (non-IID) characteristics, exacerbating data isolation problems \cite{wu2021towards}. This non-IID property can manifest as differences in node feature distributions between graph structures or clients.

Federated Learning (FL) \cite{mcmahan2017communication, SABAH2024102834,HUANG2024102576} is a distributed collaborative machine learning paradigm that offers a promising method to address the challenge of data isolation. It enables participants (i.e., clients) to train machine learning models without sharing their private data. Thus, combining FL with graph machine learning presents a promising solution to this problem \cite{10056291,HU2024102042}. However, the non-IID nature of data makes it difficult for a single global model in traditional FL methods to fit each client’s local data \cite{xie2021federated,10241965}. Additionally, data heterogeneity between clients introduces client drift in updates, leading to slower convergence \cite{karimireddy2019scaffold}. The concept of Personalized Federated Learning (PFL), which focuses on learning individualized models instead of a single global model, has been proposed to address this issue. Existing PFL methods can be classified into three types: hierarchical separation \cite{collins2021exploiting,t2020personalized}, regularization \cite{li2020federated,karimireddy2019scaffold}, and personalized aggregation \cite{luo2022adapt,huang2023fusion}. These methods explicitly or implicitly involve additional personalization parameters in their local models, with these parameters playing a dominant role in the parameter space \cite{tan2022towards}. While this addresses the problem of data heterogeneity to some extent, a significant challenge in federated graph learning is the communication cost, especially for large models. Although PFL has been extensively studied, most works focus on Euclidean data such as images. In contrast, personalized federated graph learning (PFGL) for non-Euclidean graph data is still in its early stages. It remains unclear how to effectively represent global knowledge of graphs with global components, personalize models with local components, and merge them to optimize the expressive power of local models while ensuring efficient communication.

\begin{figure}[!t]
\centering
\includegraphics[width=\textwidth]{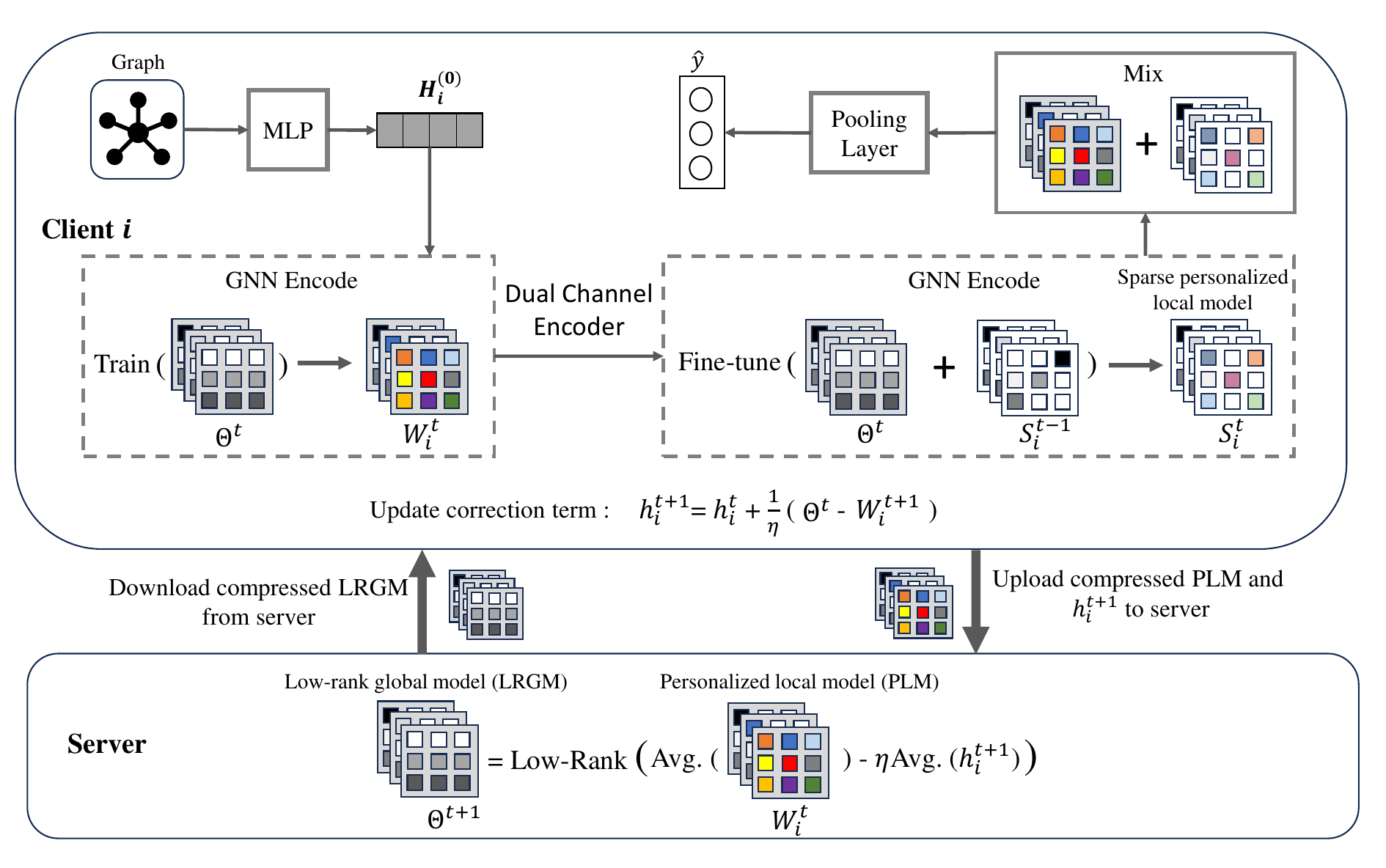}
\caption{Illustration of the \textbf{CEFGL} framework. The process begins with the client transforming the input feature data into vectors of consistent dimensions using a multilayer perceptron (MLP). The client then performs local training to obtain the initial model. During this phase, the client trains the model with its local data while keeping the model parameters \(\bm{W}_i^t\) frozen. Simultaneously, the sparsification components, which aim to reduce the complexity of the model, are fine-tuned. After local training, the client compresses the trained model using low-rank approximation techniques to reduce its size and upload it to the server. On the server side, the local gradients and the global model are updated based on the local model information received from the client. Finally, the server compresses the updated global model using low-rank methods and sends it back to the client. This process ensures efficient communication and model updates, promoting collaboration across distributed clients while minimizing communication overhead.}
\label{fig1}
\end{figure}

To tackle the challenges, we propose a communication-efficient personalized communication-efficientfederated graph learning (\textbf{CEFGL}) method, as illustrated in Figure \ref{fig1}. This method decomposes the local model into a low-rank generic model, which captures global knowledge reflecting the commonalities of clients, and a sparse private model, which comprises personalized parameters that complement the global knowledge. We employ a dual-channel encoder to learn sparse local knowledge in a personalized manner and low-rank global knowledge in a shared manner. Inspired by Scaffnew \cite{mishchenko2022proxskip}, we introduce appropriate regularization and correction terms for the low-rank and sparse components in local and global optimization problems to accelerate local training. Additionally, we utilize quantization compression to reduce the number of model parameters, reducing the communication cost for uplink and downlink. Compared to the state-of-the-art method, our proposed method demonstrates superior classification accuracy while significantly reducing communication costs. Our work has three key contributions:
\begin{itemize}
    \item We propose a communication-efficient personalized federated graph learning algorithm (\textbf{CEFGL}) for highly heterogeneous cross-dataset scenarios. The proposed method significantly enhances model performance amidst data heterogeneity while reducing communication bits and time. This highlights our unique contribution to the field, effectively addressing data heterogeneity and optimizing communication efficiency in federated graph learning.
    
    \item By utilizing low-rank decomposition techniques, our method efficiently addresses data heterogeneity. Shared knowledge across clients is encapsulated in low-rank components, while sparse components capture local-specific knowledge, resulting in significant performance improvements under non-IID data distributions. Moreover, our method integrates accelerated local training and compression techniques, substantially reducing communication costs while maintaining high performance. This dual benefit ensures high performance and efficiency, particularly effective in diverse and resource-constrained environments.
    \item Extensive experiments show that our method achieves optimal accuracy across sixteen real datasets in both IID and non-IID settings. Compared to the state-of-the-art method FedStar, our proposed method achieves a 5.64\% increase in graph classification accuracy with significantly fewer communication bits and less communication time in the cross-dataset setting CHEM (see Figure \ref{fig7}).
\end{itemize}

The rest of this paper is organized as follows. Section \ref{related} provides an overview of related work on federated learning and federated graph learning. Section \ref{pre} describes the prior knowledge used in this paper. Section \ref{method} presents the details of the proposed method. Section \ref{experiment} describes the experiments conducted to evaluate the proposed method. Finally, Section \ref{conclusion} summarizes the contributions and outlines future work.

\section{Related Work}
\label{related}
\subsection{Federated Learning}
Federated Learning (FL) methods \cite{10423871, li2020federated} aim to learn a global model by aggregating the local models of all clients. However, achieving a global model that generalizes well to each client is challenging due to data heterogeneity. Personalized Federated Learning (PFL) shifts the focus from a traditional server-centric federation to a client-centric one, better suited for non-IID data. Existing PFL methods can be categorized into three types.
The first type of method is based on regularization. Scaffold \cite{karimireddy2019scaffold} introduces proximal regularizers to mitigate client-side drift and accelerate training. FedProx \cite{li2020federated} incorporates a regularization term to balance local training with the global model, promoting faster convergence. These methods improve training and inference efficiency by controlling model parameters. However, selecting regularization parameters often requires tuning through cross-validation. Inappropriate parameter choices can over-penalise personalised parameters and lead to over-fitting of local data.
The second type involves personalized aggregation methods to learn local models. FedAMP \cite{huang2021personalized} generates aggregated models for individual clients using attention-inducing functions and personalized aggregation. APPLE \cite{luo2022adapt} performs local aggregation within each training batch rather than just local initialization. However, these methods are typically designed for cross-silo FL settings, which require all clients to participate in each iteration.
The third type separates the global and personalization layers, combining global knowledge with models personalized for each client. This paper focuses on this type of method, which is widely studied for its adaptability and learning effectiveness. For instance, FedRep \cite{collins2021exploiting} divides the backbone into a global model and client-specific headers, fine-tuning the headers locally. pFedMe \cite{t2020personalized} learns additional personalized models for clients using Moreau envelopes. FedSLR \cite{huang2023fusion} employs a two-stage proximity algorithm for optimization. Although these methods can help the model better adapt to client data, they lead to increased communication overhead as parameters are transmitted separately.

Compared to FedRep, pFedMe, and FedSLR, our method downloads a low-rank and compressed global model from the server, significantly reducing communication complexity in upstream and downstream links. We also employ accelerated local training techniques, effectively reducing communication frequency. Additionally, while the methods above are tailored for images and unsuitable for complex non-Euclidean data like graphs, our method targets highly heterogeneous cross-domain datasets. It effectively reduces communication costs while maintaining high accuracy, making it more suitable for highly heterogeneous cross-domain graph tasks.

\subsection{Federated Graph Learning}
While numerous studies have investigated federated learning, federated graph learning (FGL) remains underexplored. Unlike Euclidean data, such as images, graph data is inherently more complex, introducing new challenges and opportunities for FGL \cite{he2022spreadgnn, 9606516}. FGL enables clients to train robust GNN models in a distributed manner without sharing private data, thus expanding its application scenarios. A common strategy for training personalized federated graph learning models is client-side clustering. For instance, in IFCA \cite{ghosh2020efficient}, clients are dynamically assigned to multiple clusters based on the latest graph model gradient. However, a drawback of this method is that the clustering results are significantly influenced by the latest gradients from the clients, which are often unstable during local training. To address this issue, GCFL \cite{xie2021federated} considers a series of gradient specifications in the client cluster, and CTFL \cite{zhang2022graph} updates the global model based on representative clients for each cluster. Maintaining a shared global model in each cluster increases the number of parameters and computational requirements as the number of clients grows, leading to poor scalability. Recent approaches have focused on reducing communication costs while addressing non-IID challenges. For instance, FedStar \cite{tan2023federated} mitigates the non-IID issue by employing feature decoupling. Meanwhile, FedGCN \cite{yao2024fedgcn} reduces communication overhead by minimizing the frequency of communication with a centralized server and incorporates homomorphic encryption to further enhance privacy. Despite these advancements, these methods face challenges in highly heterogeneous cross-domain scenarios, as they can only fine-tune global model parameters to fit local data.

CEFGL significantly reduces the server's computational cost compared to clustering-based multi-model FGL methods, as it only requires generating a unified global model for all clients. Additionally, compared to FGL methods that only adjust local models, the proposed method employs a hybrid model that combines low-rank global components with sparse personalized components. This hybrid model is more adaptable to the highly heterogeneous cross-dataset graph learning task.

\section{Preliminary}
\label{pre}
\noindent \textbf{Graph Neural Networks}\quad
Let $\mathcal{G} = (\mathcal{V}, \mathcal{E})$ denote a graph, where $\mathcal{V}$ is the set of $N$ nodes, and $\mathcal{E}
$ is the set of edges. The presence of edges is represented by an adjacency matrix $\bm{A}$, where $\bm{A}_{ij} \in \{0, 1\}$ indicates the relationship between nodes $v_i$ and $v_j$ in $\mathcal{V}$. Additionally, the graph $\mathcal{G}$ is associated with a node feature matrix $\bm{X} \in \mathbb{R}^{N \times d}$, where $d$ is the feature dimension. Graph Neural Networks (GNNs) update the embedding of a given node by aggregating information from its neighbouring nodes using the following function:
\begin{equation}
\bm{X}^{(l+1)} = \text{GNN}(\bm{A}, \bm{X}^l, \bm{\omega}),
\end{equation}
where GNN($\cdot$) denotes the graph aggregation function, which can be mean, weighted average, or max/min pooling methods. Here, $\bm{X}^{(l)}$ represents the node embeddings in the $l$-th layer, and $\bm{\omega}$ is the set of trainable model weights.

For graph classification tasks, a graph-level representation $\bm{h}_{\mathcal{G}}$ can be pooled from node representations:
\begin{equation}
\bm{h}_{\mathcal{G}} = \text{readout}(\bm{X}),
\end{equation}
where readout($\cdot$) is a pooling function (e.g., mean pooling or sum pooling) that aggregates the embeddings of all nodes.

\noindent \textbf{Federated Learning}\quad In Federated Learning, we consider a set of $K$ clients $\mathcal{C} = \{c_i\}_{i=1}^K$. Each client $c_i$ owns a private dataset $\mathcal{D}_i = \{\bm{x}_j, \bm{y}_j\}_{j=1}^{N_i}$ sampled from its data distribution, where $\bm{x}_j$ is the feature vector of the $j$-th data sample and $\bm{y}_j$ is the corresponding label. Here, $N_i = |\mathcal{D}_i|$ represents the number of data samples on client $c_i$, and $N = \sum_{i=1}^K N_i$ is the total number of data samples across all clients. Let $\ell_i$ denote the loss function parameterized by $\bm{\omega}$ on client $c_i$. The goal of FL is to optimize the overall objective function:
\begin{equation}
\min_{\bm{\omega}} \sum_{i=1}^{K} \frac{N_i}{N} \mathcal{L}_{i}(\bm{\omega}) = \min_{\bm{\omega}} \frac{1}{N} \sum_{i=1}^{K} \sum_{j=1}^{N_i} \ell_j(\bm{\omega}; \bm{x}_j, \bm{y}_j),
\end{equation}
where $\mathcal{L}_i$ is the average loss over the local data on client $c_i$.

\begin{algorithm*}[t]
   \caption{CEFGL}
   \label{alg1}
   \textbf{Input}: $K$ clients, client joining ratio $\rho$, global epochs $T$, local epochs $E$, fine-tune epochs $F$, probability $p$, stepsize $\eta$, 
  Data $\mathcal{D}_{i}$, compressor $Q_r(\cdot)$ \\
   \begin{algorithmic}[1]
      \STATE Initialize client model $\bm{W}_i^0 =\bm{\Theta}^0$.
      \FOR{$t=0$ {\bfseries to} $T-1$}
      \STATE $p$: Server flips a coin, where $\text{Prob}(j = 1) = p$, $0 \leq p \leq 1, j \in \{0, 1\}$ \hfill $\lozenge$ \hspace{0.1em} \textit{Decide when to skip communication}
      \STATE $\mathcal{C}_{\rho}$: Server randomly selects $K \times \rho$ clients
      \FOR{Client $c_i$ $\in$ $\mathcal{C}_{\rho}$}
      \STATE Client update model: $\bm{W}_i^{t+1} = \bm{W}_i^{t}-\eta(\nabla_{\bm{W}}f_i(\bm{W}_i^{t};\mathcal{D}_{i})-h_i^{t})+\frac{\alpha}{2}\|\bm{\Theta}^t-\bm{W}_i^t\|$ \hfill $\lozenge$ \hspace{0.1em} \textit{Train low-rank local models}
      
      \STATE Client update sparse model: $\bm{S}_i^{t+1} = \operatorname{Sparse}(\bm{S}_i^{t}-\eta\nabla_{\bm{S}}f_i(\bm{\Theta}^{t}+\bm{S}_i^{t};\mathcal{D}_{i}) + \|\bm{S}^t_i\|)$ \hfill $\lozenge$ \hspace{0.1em} \textit{Fine-tune sparse models}
      \STATE Update the correction term: $\bm{h}_{i}^{t+1}=\bm{h}_{i}^{t}+\frac{1}{\eta}(\bm{\Theta}^t-\bm{W}_i^{t+1})$ \hfill $\lozenge$ \hspace{0.1em} \textit{Update the correction term $\bm{h}_{i}^{t}$}
      \STATE Quantitative compression: $\bm{W}_i^{t+1} =Q_r(\bm{W}_i^{t+1})$; $\bm{h}_{i}^{t+1} = Q_r(\bm{h}_{i}^{t+1})$ \hfill $\lozenge$ \hspace{0.1em} \textit{Uplink compression}
      \ENDFOR

      \IF{$j=1$}
      \STATE Server aggregates local models: 
      $\bm{\Theta}^{t+1} = Q_r(\operatorname{Low-rank} ( \sum_{i=1}^{|\mathcal{C}_{\rho}|}\frac{|\mathcal{D}_i|}{|\mathcal{D}|}\bm{W}_i^{t+1} - \eta\sum_{i=1}^{|\mathcal{C}_{\rho}|}\frac{|\mathcal{D}_i|}{|\mathcal{D}|} \bm{h}_{i}^{t+1}))$
      \ELSE
      \STATE $\bm{\Theta}^{t+1} = \bm{W}_{i}^{t+1}$. \hfill $\lozenge$ \hspace{0.1em} \textit{Skip communication}
      \ENDIF
      \ENDFOR
   \end{algorithmic}
\end{algorithm*}

\section{Methodology}
\label{method}
Based on the previous section, we propose a communication-efficient personalized federated graph learning algorithm (\textbf{CEFGL}). Drawing inspiration from RPCA \cite{candes2011robust}, we decompose client model parameters into low-rank and sparse components. The low-rank component captures common knowledge among clients, while the sparse component captures unique, client-specific knowledge. To reduce communication costs and computational overhead, we integrate compression techniques and perform low-rank approximation only during the global aggregation phase. The overall pseudo-code for CEFGL is presented in Algorithm \ref{alg1}.

\subsection{Training and Fine-tuning on the Client Side}
In robust principal component analysis (RPCA) \cite{candes2011robust}, the matrix $\bm{M}$ is decomposed into a low-rank matrix $\bm{W}$ and a sparse matrix $\bm{S}$, where $\bm{W}$ captures the principal components retaining the maximum information in $\bm{M}$, while $\bm{S}$ represents the sparse outliers. Motivated by RPCA, we decompose the local model into a shared low-rank component and a private sparse component, expressed as:
\begin{equation}
\underset{\bm{W},\bm{S}}{\operatorname*{\operatorname{argmin}}}rank(\bm{W})+\nu\|\mathbf{S}\|_0\quad s.t.\quad\bm{M}=\bm{W}+\bm{S},
\end{equation}
where $rank(\bm{W})$ is the rank of the matrix $\bm{W}$, $\|\bm{S}\|_0$ is the number of nonzero elements in $\bm{S}$, and $\nu$ is the parameter used to trade-off between these two terms. Since the rank function and the $L_0$-norm are non-convex, we can minimize their convex proxies as
\begin{equation}
\underset{\bm{W},\bm{S}}{\operatorname*{\operatorname{argmin}}}\|\bm{W}\|_*+\nu\|\mathbf{S}\|_1\quad s.t.\quad\bm{M}=\bm{W}+\bm{S},
\end{equation}
where $\|\bm{W}\|_*$ is the trace paradigm number of $\bm{W}$, i.e., the sum of the singular values of $\bm{W}$. $\|\bm{S}\|_1$ is the $L_1$-norm, i.e., the sum of the absolute values of all elements in $\bm{S}$.

\textbf{Local training}\quad In the local training phase, we first use a multilayer perceptron (MLP) to transform feature information into vectors of consistent dimensions. Subsequently, we employ a dual-channel encoder to represent the low-rank component of global knowledge and the sparse component of personalized knowledge. Inspired by Scaffnew \cite{mishchenko2022proxskip}, we introduce a correction term \( \bm{h}_i^t \) to record the local gradient, accelerating local training and mitigating client drift. This term contains information about the client's model updating direction, akin to the client-drift value of the global model relative to the local model. The optimization process for learning the low-rank component is expressed as follows:
\begin{equation}
\bm{W}_i^{t+1} = \bm{W}_i^{t}-\eta(\nabla_{\bm{W}}f_i(\bm{W}_i^{t};\mathcal{D}_{i})-h_i^{t})+\frac{\alpha}{2}\|\bm{\Theta}^t-\bm{W}_i^t\|,
\label{eq7}
\end{equation}
where $\eta$ represents the learning rate. $\alpha$ is a hyper-parameter used to balance the weight between the global model and the prior local model. $\bm{\Theta}^t$ is the global model downloaded from the server in the $t$-th iteration, and $\bm{W}_i^t$ is the low-rank component on client $i$ at the $t$-th iteration. The function $f$ represents the neural network transformation, and the cross-entropy loss function is employed. Initially, $\bm{W}_i^{0} = \bm{\Theta}^0$.

After completing the local training, we update the correction term in each round as follows:
\begin{equation}
    \bm{h}_{i}^{t+1}=\bm{h}_{i}^{t}+\frac{1}{\eta}(\bm{\Theta}^t-\bm{W}_i^{t+1}).
\label{eq8}
\end{equation}

\textbf{Fine-tuning}\quad In the fine-tuning phase, we freeze the low-rank component $\bm{W}_i^t$ learned in the local training phase and fine-tune the sparse component using the initialized global parameters to fuse personalized knowledge with global knowledge. The optimization process of sparse component learning can be represented as follows:
\begin{equation}
    \bm{S}_i^{t+1} = \operatorname{Sparse}(\bm{S}_i^{t}-\eta\nabla_{\bm{S}}f_i(\bm{\Theta}^{t}+\bm{S}_i^{t};\mathcal{D}_{i}) + \|\bm{S}^t_i\|_1),
\label{eq9}
\end{equation}
where $\operatorname{Sparse}(\cdot)$ represents the sparsification function. We explore two types of sparsification: threshold sparsification and Top-$k$ sparsification. Threshold sparsification involves setting model parameters with absolute values smaller than a given threshold $\mu$ to zero, thus achieving parameter sparsity. Top-$k$ sparsification retains the \( k \) parameters with the largest absolute values in the parameter matrix, setting the rest to zero.
\begin{remark}
To reduce the computational cost, we defer the intensive computation of low-rank decomposition to the aggregation phase, where the server performs it, typically having more computational resources. Although the parameter \( \bm{W}_i^t \) is not low-rank during local training, we refer to it as the low-rank component for clarity. In the fine-tuning phase, we optimize the sparse component \( \bm{S}_i^t \) using stochastic gradient descent to integrate personalized knowledge. The sparse component, with few parameters, does not dominate the local model or cause overfitting. Despite applying \( L_1 \) regularization in each local fine-tuning step, the overhead remains insignificant compared to low-rank regularization.
\end{remark}

\subsection{Aggregation on the Server Side}
After completing the local training and fine-tuning, the client sends the trained low-rank component $\bm{W}_i^t$ and the updated correction term $\bm{h}_i^{t+1}$ to the server. The server aggregates the client models and performs the low-rank decomposition, which can be represented as follows:
\begin{equation}
 \bm{\Theta}_{i}^{t+1} =\operatorname{Low-rank} \Big( \sum_{i=1}^{K}\frac{|\mathcal{D}_i|}{|\mathcal{D}|}\bm{W}_{i+1}^t - \eta\sum_{i=1}^{K}\frac{|\mathcal{D}_i|}{|\mathcal{D}|} \bm{h}_{i}^{t+1}\Big),
\label{eq10}
\end{equation}
where $K$ denotes the number of clients participating in training, $|\mathcal{D}_i|$ denotes the total number of samples on client $c_i$, $|\mathcal{D}|$ denotes the total number of samples on $K$ clients, and $\operatorname{Low-rank}(\cdot)$ denotes the low-rank regularization. In this paper, the low-rank method we use is truncated singular value decomposition (TSVD), i.e., only singular values larger than a specific threshold $\lambda$ and their corresponding vectors are retained.
\begin{remark}
During local training, when Eq. \eqref{eq7} reaches the optimal solution, we have \( \bm{h}_i^t = \nabla_{\bm{W}} f_i(\bm{W}_i^{t}) \). We substitute this relationship into Eq. \eqref{eq10}. If the local training converges, i.e., \( \bm{W}_i^{t+1} = \bm{\Theta}^t \), then \( \nabla_{\bm{W}} f_i(\bm{W}_i^{t}) \) can be approximated as \( \nabla_{\bm{\Theta}} f_i(\bm{\Theta}^{t}) \). At this point, the optimization problem for global aggregation can be expressed as \( \bm{\Theta}^{t+1} = \operatorname{Low-rank}(\bm{\Theta}^t - \eta \nabla_{\bm{\Theta}} f_i(\bm{\Theta}^{t})) \). This update essentially performs an unbiased gradient descent update of the loss function.
\end{remark}

\subsection{Quantization Compression}
In the algorithm described above, the model after server aggregation is low-rank, reducing communication costs to some extent. However, since the local model uploaded by the client remains dense, the uplink communication cost is not reduced. To address this, we adopt a strategy of updating the local parameters several times before aggregation and integrating quantization compression into the algorithm.

Quantization compression involves converting model parameters from raw floating-point values to representations using fewer bits. While this reduces the accuracy of parameter representation, it significantly reduces the storage space and memory footprint of the model, thereby enhancing its efficiency during inference. For any vector \( \bm{x} \in \mathbb{R}^d \), with \( \bm{x} \neq 0 \) and a number of bits \( r > 0 \), its binary quantization \( Q_r \) is defined componentwise as follows:

\begin{equation}
Q_r\left(\bm{x}\right)=\left(\left\|\bm{x}\right\|_2\cdot\operatorname{sgn}\left(\bm{x}_i\right)\cdot\xi_i(\bm{x},2^r)\right)_{1\leq i\leq d},
\end{equation}
where $\xi_i(\bm{x},2^r)$ is independent random variable and $\|\cdot\|_2$ is $L_2$-norm. The probability distribution is given by
\begin{equation}
\xi_i(\bm{x},2^r)=\operatorname{round}\left(\frac{2^r|\bm{x}_i|}{\|\bm{x}\|_2}\right)/2^r,
\end{equation}
where round($\cdot$) means round to the nearest integer.

Our proposed algorithm reduces the need for frequent client-server communication, significantly lowering communication frequency. Importantly, our method is less susceptible to data leakage than other methods, as the uncertain local step size prevents the server from accurately recovering the client's actual gradient. Additionally, clients use quantization compression to reduce parameter size from 32 bits to 4 bits before uploading them to the server. Similarly, the server compresses the model before sending it to the client. This compression is particularly beneficial for downloading the model from the server, which can be costly due to network bandwidth limitations. Overall, these strategies significantly reduce both uplink and downlink communication costs.

\begin{remark}
The proposed method has several distinct advantages. First, our dual-channel encoder effectively captures global and personalized knowledge, ensuring that the model benefits from shared insights and client-specific information. In addition, introducing correction terms helps mitigate client-side drift and speeds up local training, thereby improving the overall convergence speed. By deferring the intensive computation of low-rank decomposition to the server, we leverage the superior computational resources of the server and significantly reduce the computational burden on clients.

Another highlight of CEFGL is the integration of quantization compression with multi-step local training. This significantly reduces the communication cost and enhances data security by preventing the server from accurately recovering the actual gradient from the client due to uncertainty in the local step size. These combined strategies make our method efficient, secure, and practical, standing out among existing federated learning methods.
\end{remark}

\section{Experimental Evaluation}
\label{experiment}
\subsection{Experimental Setup}
\noindent \textbf{Datasets}\quad We utilize sixteen public graph classification datasets from four domains\footnote{https://chrsmrrs.github.io/datasets/docs/datasets}. These include seven datasets from the small molecule domain (MUTAG, BZR, COX2, DHFR, PTC MR, AIDS, NCI1), three from bioinformatics (ENZYMES, DD, PROTEINS), three from social networking (COLLAB, IMDB-BINARY, IMDB-MULTI), and three from computer vision (Letter-low, Letter-high, Letter-med). To evaluate the performance of the proposed method, we adopt the same setup as GCFL \cite{xie2021federated}, creating three IID and six non-IID setups. The IID setups consist of three single datasets (DD, NCI1, IMDB-MULTI). The non-IID setups consist of (1) cross-dataset setups based on the seven small molecule datasets (CHEM) and (2)-(5) setups based on data from two or three domains (SN\_CV, CHEM\_BIO, CHEM\_BIO\_SN, BIO\_SN\_CV) in cross-domain settings, along with (6) a cross-domain setup encompassing all datasets (ALL).

\noindent \textbf{Baselines}\quad We compare CEFGL with seven benchmark methods. These include (1) the standard FL algorithm FedAvg \cite{mcmahan2017communication}; (2)-(3) some existing PFL solutions, namely FedProx \cite{li2020federated} and FedPer \cite{arivazhagan2019federated}; and (4)-(7) four state-of-the-art FGL methods: GCFL \cite{xie2021federated}, SpreadGNN \cite{he2022spreadgnn}, FedGCN \cite{yao2024fedgcn}, and FedStar \cite{tan2022towards}. We use the same experimental setup and data partitioning criteria for all methods to ensure fair comparisons and tune the hyperparameters to their best states.

\noindent \textbf{Default Configuration}\quad For CEFGL, we employ three two-layer graph neural networks (GCN \cite{kipf2016semi}, GIN \cite{XuHLJ19}, GraphSage \cite{HamiltonYL17}) as two-channel encoders to train the low-rank component (\( \bm{W}_i^t \)) and the sparse component (\( \bm{S}_i^t \)). We set the low-rank hyperparameter \( \mu \) to 0.0001, the sparse hyperparameter \( \lambda \) to 0.001, and the communication probability \( p \) to 0.5. During training, we set the hyperparameters \(\nu\) in Eq. (4) and \(\alpha\) in Eq. (6) to 0.5 and 0.6, respectively. The dataset is randomly split into training (80\%), validation (10\%), and test sets (10\%). We divide the dataset into ten clients using the same criteria in the single dataset setup. For cross-dataset setups, each dataset is assigned to one client. We conduct 200 rounds of global training with consistent data partitioning across all methods. The Adam optimizer has a weight decay of 5e-4 and a learning rate of 0.001. We average the results over five runs with different random seeds to ensure fairness, reporting the average accuracy and corresponding standard deviation. All experiments use PyTorch on a server with four NVIDIA 3090 GPUs (with 24G memory).

\begin{table}[]
\renewcommand{\arraystretch}{1.15}
\caption{Accuracy (\%) on six cross-datasets non-IID settings for the graph classification. The best result is bold, and the second best is underlined. The last column shows the average classification accuracy of the method across all nine data settings.\label{tab1}}
\centering
\begin{adjustbox}{max width=\textwidth}
\begin{tabular}{ccccccccc}
\hline
\multicolumn{2}{c}{\multirow{2}{*}{Method}}                 & \multicolumn{7}{c}{Dataset}                                                                                                                                            \\ \cline{3-9} 
\multicolumn{2}{c}{}                                        & CHEM           & SN\_CV         & CHEM\_BIO      & CHEM\_BIO\_SN  & BIO\_SN\_CV    & ALL            &Avg. acc           \\ \hline
\multicolumn{1}{c|}{\multirow{9}{*}{GCN}}      & FedAvg    & 78.64 $\pm$ 0.35          & 70.17 $\pm$ 0.26          & 71.97 $\pm$ 0.18          & 71.99 $\pm$ 0.50      & 68.97 $\pm$ 0.41          & 70.91 $\pm$ 0.27            & 72.11        \\
\multicolumn{1}{c|}{}                           & FedProx   & 77.20 $\pm$ 0.38          & 63.30 $\pm$ 0.42          & 72.92 $\pm$ 0.29          & 71.43 $\pm$ 0.52          & 61.40 $\pm$ 0.25          & 62.38 $\pm$ 0.60           & 68.11          \\
\multicolumn{1}{c|}{}                           & FedPer    & 78.45 $\pm$ 0.36          & 70.80 $\pm$ 0.50          & 74.17 $\pm$ 0.47          & 73.35 $\pm$ 0.48          & 69.27 $\pm$ 0.52          & 71.89 $\pm$ 0.40          & 72.99          \\
\multicolumn{1}{c|}{}                           & GCFL      & 78.41 $\pm$ 0.21          & 70.95 $\pm$ 0.53          & 73.72 $\pm$ 0.20          & 72.98 $\pm$ 0.44          & 69.10 $\pm$ 0.50          & 71.54 $\pm$ 0.47          & 72.78          \\
\multicolumn{1}{c|}{}                           & SpreadGNN & 79.85 $\pm$ 0.43          & 72.53 $\pm$ 0.45          & 74.07 $\pm$ 0.40          & 72.28 $\pm$ 0.40          & 70.54 $\pm$ 0.62          & 74.31 $\pm$ 0.25          & 73.93          \\
\multicolumn{1}{c|}{}                           & FedGCN & 80.47 $\pm$ 0.53          & 73.33 $\pm$ 0.30          & {\ul74.61 $\pm$ 0.45} & 73.27 $\pm$ 0.47          & 71.72 $\pm$ 0.52          & 73.99 $\pm$ 0.26     & 74.56    \\
\multicolumn{1}{c|}{}                           & FedStar   & {\ul 82.23 $\pm$ 0.35}    & {\ul 73.53 $\pm$ 0.43}    & 74.28 $\pm$ 0.45          & {\ul 73.49 $\pm$ 0.37}    & {\ul 71.78 $\pm$ 0.35}    & {\ul 74.41 $\pm$ 0.44}              & {\ul 74.95}          \\
\multicolumn{1}{c|}{}                           & Ours      & \textbf{85.81 $\pm$ 0.38} & \textbf{74.14 $\pm$ 0.45} & \textbf{77.57 $\pm$ 0.47}    & \textbf{73.68 $\pm$ 0.28} & \textbf{72.90 $\pm$ 0.43}  & \textbf{74.83 $\pm$ 0.39}  & \textbf{76.40} \\ \hline
\multicolumn{1}{c|}{\multirow{9}{*}{GIN}}       & FedAvg    & 76.82 $\pm$ 0.25          & 67.29 $\pm$ 0.41          & 69.46 $\pm$ 0.43          & 70.02 $\pm$ 0.35          & 63.81 $\pm$ 0.52          & 67.67 $\pm$ 0.48          & 69.18          \\
\multicolumn{1}{c|}{}                           & FedProx   & 74.99 $\pm$ 0.27          & 47.46 $\pm$ 0.50          & 68.56 $\pm$ 0.35          & 67.89 $\pm$ 0.29          & 47.53 $\pm$ 0.46          & 61.00 $\pm$ 0.52          & 61.24          \\
\multicolumn{1}{c|}{}                           & FedPer    & 73.91 $\pm$ 0.42          & 72.31 $\pm$ 0.50          & 70.04 $\pm$ 0.43          & 73.94 $\pm$ 0.26          & 64.32 $\pm$ 0.65          & 73.11 $\pm$ 0.47          & 71.27          \\
\multicolumn{1}{c|}{}                           & GCFL      & 75.58 $\pm$ 0.53          & 68.66 $\pm$ 0.48          & 69.57 $\pm$ 0.41          & 66.50 $\pm$ 0.16          & 59.24 $\pm$ 0.28          & 68.07 $\pm$ 0.12          & 67.94          \\
\multicolumn{1}{c|}{}                           & SpreadGNN & 80.01 $\pm$ 0.09          & 72.87 $\pm$ 0.29          & 74.31 $\pm$ 0.36          & 72.13 $\pm$ 0.42          & 70.70 $\pm$ 0.47          & 74.06 $\pm$ 0.50          & 74.01          \\
\multicolumn{1}{c|}{}                           & FedGCN & {\ul 80.92 $\pm$ 0.25}    & 72.97 $\pm$ 0.39          & 74.52 $\pm$ 0.56          & {\ul 73.98 $\pm$ 0.37}    & {\ul 71.60 $\pm$ 0.65}    & 74.37 $\pm$ 0.46    & {\ul 74.72}    \\
\multicolumn{1}{c|}{}    & FedStar   & 79.79 $\pm$ 0.38          & 73.26 $\pm$ 0.36          & {\ul 74.54 $\pm$ 0.57}    & 72.16 $\pm$ 0.55          & 69.49 $\pm$ 0.48          & {\ul 74.45 $\pm$ 0.43}        & 73.94          \\
\multicolumn{1}{c|}{}                           & Ours      & \textbf{85.43 $\pm$ 1.14} & \textbf{74.21 $\pm$ 0.52} & \textbf{78.82 $\pm$ 0.47} & \textbf{74.13 $\pm$ 0.38} & \textbf{73.35 $\pm$ 0.50} & \textbf{75.05 $\pm$ 0.33} & \textbf{76.83} \\ \hline
\multicolumn{1}{c|}{\multirow{9}{*}{GraphSage}}  & FedAvg    & 79.13 $\pm$ 0.35          & 72.17 $\pm$ 0.42          & 74.08 $\pm$ 0.45          & 73.10 $\pm$ 0.26          & 72.11 $\pm$ 0.51          & 73.89 $\pm$ 0.42          & 74.08          \\
\multicolumn{1}{c|}{}                           & FedProx   & 70.45 $\pm$ 0.33          & 57.54 $\pm$ 0.41          & 67.11 $\pm$ 0.19          & 69.72 $\pm$ 0.20          & 56.79 $\pm$ 0.29          & 61.09 $\pm$ 0.45         & 63.78          \\
\multicolumn{1}{c|}{}                           & FedPer    & 79.49 $\pm$ 0.31          & 73.31 $\pm$ 0.27          & 74.53 $\pm$ 0.46          & 73.03 $\pm$ 0.54          & 72.35 $\pm$ 0.43          & 73.94 $\pm$ 0.30          & 74.44          \\
\multicolumn{1}{c|}{}                           & GCFL      & 79.86 $\pm$ 0.42          & 72.87 $\pm$ 0.26          & 74.35 $\pm$ 0.18          & 73.08 $\pm$ 0.36          & {\ul 72.70 $\pm$ 0.21}  & 73.90 $\pm$ 0.18   & 74.46          \\
\multicolumn{1}{c|}{}                           & SpreadGNN & 81.14 $\pm$ 0.21          & 73.17 $\pm$ 0.24          & 73.70 $\pm$ 0.40          & 72.54 $\pm$ 0.23          & 71.01 $\pm$ 0.23          & 74.13 $\pm$ 0.33          & 74.28          \\
\multicolumn{1}{c|}{}                           & FedGCN & 80.57 $\pm$ 0.40          & 73.07 $\pm$ 0.25          & 74.16 $\pm$ 0.23          & 73.15 $\pm$ 0.25    & 70.59 $\pm$ 0.52          & 74.08 $\pm$ 0.22          & 74.27   \\
\multicolumn{1}{c|}{}                           & FedStar   & {\ul 81.71 $\pm$ 0.42}    & {\ul 73.36 $\pm$ 0.29}    &{\ul 75.13 $\pm$ 0.30} & {\ul 74.25 $\pm$ 0.25}    & 72.18 $\pm$ 0.25          & {\ul 74.88 $\pm$ 0.40}    & {\ul 75.24}          \\
\multicolumn{1}{c|}{}                           & Ours      & \textbf{85.15 $\pm$ 0.51} & \textbf{74.67 $\pm$ 0.27} &\textbf{78.45 $\pm$ 0.32}    & \textbf{74.57 $\pm$ 0.40} & \textbf{73.56 $\pm$ 0.26}    & \textbf{75.69 $\pm$ 0.57} & \textbf{77.02} \\ \hline
\end{tabular}
\end{adjustbox}
\end{table}

\begin{table}[t]
\renewcommand{\arraystretch}{1}
\caption{Accuracy (\%) on three single-datasets for the graph classification. The best result is bold, and the second best is underlined. The last column shows the average classification accuracy of the method across all nine data settings.\label{tab11}}
\centering
\begin{adjustbox}{max width= 0.9\textwidth}{
\begin{tabular}{cccccc}
\hline
\multicolumn{2}{c}{\multirow{2}{*}{Method}}                 & \multicolumn{4}{c}{Dataset}                                                                                                                                            \\ \cline{3-6} 
\multicolumn{2}{c}{}    & DD             & NCI1           & IMDB-BINARY    & Avg. acc           \\ \hline
\multicolumn{1}{c|}{\multirow{9}{*}{GCN}}      & FedAvg    & 71.56 $\pm$ 1.02          & 66.84 $\pm$ 0.54          & 70.31 $\pm$ 0.45          & 69.31        \\
\multicolumn{1}{c|}{}                           & FedProx   & 72.66 $\pm$ 0.45          & 65.63 $\pm$ 0.60          & 69.53 $\pm$ 0.38          & 69.27          \\
\multicolumn{1}{c|}{}                           & FedPer    & 74.20 $\pm$ 0.43          & 66.40 $\pm$ 0.41          & 67.97 $\pm$ 0.50          & 69.53          \\
\multicolumn{1}{c|}{}                           & GCFL      & {\ul 74.22 $\pm$ 0.38}    & 65.63 $\pm$ 0.45         & 67.96 $\pm$ 0.23          & 69.27          \\
\multicolumn{1}{c|}{}                           & SpreadGNN & 69.35 $\pm$ 0.36          &{\ul 68.34 $\pm$ 0.44}  & {\ul 73.28 $\pm$ 0.38}    & {\ul 70.32}          \\
\multicolumn{1}{c|}{}                           & FedGCN & 69.55 $\pm$ 0.54          & 67.56 $\pm$ 0.37          & 71.04 $\pm$ 0.44          & 69.38    \\
\multicolumn{1}{c|}{}                           & FedStar   & 66.41 $\pm$ 0.42          & 66.87 $\pm$ 0.53          & 72.85 $\pm$ 0.48          & 68.71          \\
\multicolumn{1}{c|}{}                           & Ours      & \textbf{74.31 $\pm$ 0.40} & \textbf{68.84 $\pm$ 0.42}    & \textbf{75.15 $\pm$ 0.34} & \textbf{72.77} \\ \hline
\multicolumn{1}{c|}{\multirow{9}{*}{GIN}}       & FedAvg    & 67.94 $\pm$ 1.30          & 66.10 $\pm$ 0.50          & 70.93 $\pm$ 0.28          & 68.32          \\
\multicolumn{1}{c|}{}                           & FedProx   & 66.59 $\pm$ 0.50          & 64.18 $\pm$ 0.47          & 73.09 $\pm$ 0.36          & 67.95          \\
\multicolumn{1}{c|}{}                           & FedPer    & 66.87 $\pm$ 0.53          & 66.15 $\pm$ 0.25          & 73.16 $\pm$ 0.28          & 68.73          \\
\multicolumn{1}{c|}{}                           & GCFL      & 66.94 $\pm$ 0.35          & 66.11 $\pm$ 0.53          & 73.23 $\pm$ 0.46          & 68.76          \\
\multicolumn{1}{c|}{}                           & SpreadGNN  & 68.62 $\pm$ 0.36          & 67.31 $\pm$ 0.50          & {\ul 73.85 $\pm$ 0.47}    & 69.93          \\
\multicolumn{1}{c|}{}                           & FedGCN & 69.31 $\pm$ 0.42          & 67.28 $\pm$ 0.51          & 73.49 $\pm$ 0.47          & 70.03    \\
\multicolumn{1}{c|}{}                           & FedStar   & {\ul 71.76 $\pm$ 0.28}    & {\ul 68.71 $\pm$ 0.43}    & 73.67 $\pm$ 0.29          & {\ul 71.38}          \\
\multicolumn{1}{c|}{}                           & Ours      & \textbf{75.96 $\pm$ 0.38} & \textbf{69.94 $\pm$ 0.62} & \textbf{75.43 $\pm$ 0.27} & \textbf{73.78} \\ \hline
\multicolumn{1}{c|}{\multirow{9}{*}{GraphSage}}  & FedAvg    & 70.31 $\pm$ 0.35          & 63.28 $\pm$ 0.28          & 69.53 $\pm$ 0.21          & 67.71          \\
\multicolumn{1}{c|}{}                           & FedProx   & 67.97 $\pm$ 0.36          & {\ul 67.86  $\pm$ 0.51}    & 66.47 $\pm$ 0.27          & 67.43          \\
\multicolumn{1}{c|}{}                           & FedPer    & 66.52 $\pm$ 0.32          & 65.63 $\pm$ 0.40          & 71.88 $\pm$ 0.19          & 68.01          \\
\multicolumn{1}{c|}{}                           & GCFL      & 69.53 $\pm$ 0.29          & 67.71 $\pm$ 0.31          & 67.97 $\pm$ 0.20          & 68.40          \\
\multicolumn{1}{c|}{}                           & SpreadGNN & 69.23 $\pm$ 0.60          & 67.74 $\pm$ 0.38          & 73.46 $\pm$ 0.42          & 70.14          \\
\multicolumn{1}{c|}{}                           & FedGCN & {\ul 70.43 $\pm$ 0.50}    & 67.36 $\pm$ 0.26          & {\ul 73.87 $\pm$ 0.23}    & {\ul 70.55}    \\
\multicolumn{1}{c|}{}                           & FedStar   & 64.84 $\pm$ 0.33          & 67.31 $\pm$ 0.36          & 70.31 $\pm$ 0.26          & 67.49          \\
\multicolumn{1}{c|}{}                           & Ours      & \textbf{74.56 $\pm$ 0.44} & \textbf{68.07 $\pm$ 0.45} & \textbf{75.08 $\pm$ 0.20} & \textbf{72.57} \\ \hline
\end{tabular}}
\end{adjustbox}
\end{table}

\subsection{Performance Evaluation}
To comprehensively evaluate the performance of CEFGL, we conducted experiments using three commonly used graph neural network architectures: GIN, GCN, and GraphSage. TABLE \ref{tab1} and TABLE \ref{tab11} present the average test accuracy and standard deviation for graph classification across six non-IID cross-dataset and three single-dataset settings. Our results consistently demonstrate that CEFGL outperforms all baseline methods across the three GNN architectures. In the highly heterogeneous cross-dataset graph learning task, CEFGL achieves an accuracy that is 2.11\% higher than FedPerGCN,  2.82\% higher than SpreadGNN, and 6.92\% higher than GCFL. This significant improvement highlights the ability of CEFGL to effectively learn and generalize across diverse datasets, leveraging its dual-channel encoder to capture global and personalized knowledge efficiently. In the single dataset task, some personalization solutions exhibit performance degradation in the IID setting, failing to match the performance of global solutions like FedAvg. However, CEFGL maintains the highest performance in the IID setting, which indicates that its sparse personalization information effectively complements global information without overwhelming it, thereby preserving model generalization. This balance between personalization and generalization is crucial for maintaining high performance across different data distributions.

The superiority of CEFGL is further highlighted by its performance consistency across various setups. In non-IID settings, where data heterogeneity poses significant challenges, the architecture of CEFGL ensures effective communication and learning, reducing client drift and enhancing convergence. In IID settings, the ability of CEFGL to retain high accuracy showcases its adaptability and strength in more uniform data distributions. These comprehensive results underscore the effectiveness of CEFGL in both heterogeneous non-IID and IID settings. By consistently outperforming existing methods, CEFGL proves its superiority in federated graph learning, offering a reliable solution for real-world applications where data distribution can vary significantly.

\subsection{Low-rank Ratios Analysis}
To evaluate the impact of the low-rank hyperparameter \( \mu \) on the performance of the algorithm, we vary \( \mu \) and record the low-rank rate and classification accuracy of the model \( \bm{\Theta}^t \) for different values of \( \mu \) using GIN as the backbone, as shown in TABLE \ref{tab2}. A \( \mu \) value 0 indicates no low-rank regularization, resulting in dense models susceptible to non-IID data from other clients, leading to suboptimal performance in federated environments with heterogeneous data. As \( \mu \) increases from 0, we observe an initial improvement in classification accuracy and significant reductions in communication cost and the number of parameters. This is because low-rank regularization helps filter out noise, capture common patterns in the data, and enhance the generalization of models across clients. For instance, the model balances compactness and representativeness at \( \mu = 0.001 \), improving communication efficiency and accuracy. This optimal setting maximizes the extraction of global knowledge while maintaining a manageable number of parameters. However, if $\mu$ is set too large, for instance, 0.01, the model becomes overly constrained by the low-rank regularization. This excessive constraint can cause the model to lose important information for accurate classification, resulting in significant performance degradation. In conclusion, the low-rank hyperparameter \( \mu \) plays a crucial role in balancing the trade-off between communication cost and classification accuracy. By choosing an appropriate \( \mu \), our method enhances the extraction of global knowledge and improves overall model performance, demonstrating the effectiveness of using low-rank regularization in federated graph learning.

\begin{table}[t]
\renewcommand{\arraystretch}{1}
\caption{Parameter sensitivity of low-rank hyperparameter $\mu$.}
\label{tab2}
\centering
\resizebox{0.9\textwidth}{!}{
\begin{tabular}{ccccccc}
\hline
                        & Low-rank parameter $\mu$ ($\lambda$=0.0001) & 0     &0.0001 
& 0.0005 & 0.001  
 &0.01\\ \hline
\multirow{2}{*}{DD}     & low-rank ratio               & 100\% &98\%   
& 92\%   & 85\%   
 &68\%\\
                        & accuracy (\%)                     & 71.30  &74.15  
& 74.15  & \textbf{75.96}  
 &74.87\\ \hline
\multirow{2}{*}{CHEM}   & low-rank ratio               & 100\% &98\%   
& 92\%   & 87\%   
 &71\%\\
                        & accuracy (\%)                     & 83.79 &84.54  
& 84.61  & \textbf{85.53}  
 &85.21\\ \hline
\multirow{2}{*}{SN\_CV} & low-rank ratio               & 100\% &98\%   
& 94\%   & 88\%   
 &75\%\\
                        & accuracy (\%)                      & 74.06 &73.79  & \textbf{74.3}   & 74.21   &74.14\\ \hline
\end{tabular}}
\end{table}

\begin{table}[t]
\renewcommand{\arraystretch}{1}
\caption{Parameter sensitivity of sparse hyperparameter $\lambda$.}
\label{tab3}
\centering
\resizebox{0.85\textwidth}{!}{
\begin{tabular}{ccccccc}
\hline
\textbf{}               & Sparse parameter $\lambda$ ($\mu$=0.001) & 0     & 0.00005 
&0.0001 
& 0.0005 & 0.001  
\\ \hline
\multirow{2}{*}{DD}     & density ratio             & 100\% & 
78\%&71\%   
& 19\%   & 4\%    
\\
                        & accuracy(\%)              & 72.44 & 72.46&\textbf{75.96}  
& 73.51  &73.87  
\\ \hline
\multirow{2}{*}{CHEM}   & density ratio             & 100\% & 
69\%&54\%   
& 12\%   & 8\%    
\\
                        & accuracy(\%)              & 83.47 & 83.81&84.55  
& 84.8   & \textbf{85.03} 
\\ \hline
\multirow{2}{*}{SN\_CV} & density ratio             & 100\% & 
61\%&43\%   
& 6\%    & 2\%    
\\
                        & accuracy(\%)              & 74.15 & 73.40&74.50& \textbf{74.52}& 74.21  \\ \hline
\end{tabular}}
\end{table}

\begin{figure}[!t]
\centering
\includegraphics[width=0.4\textwidth]{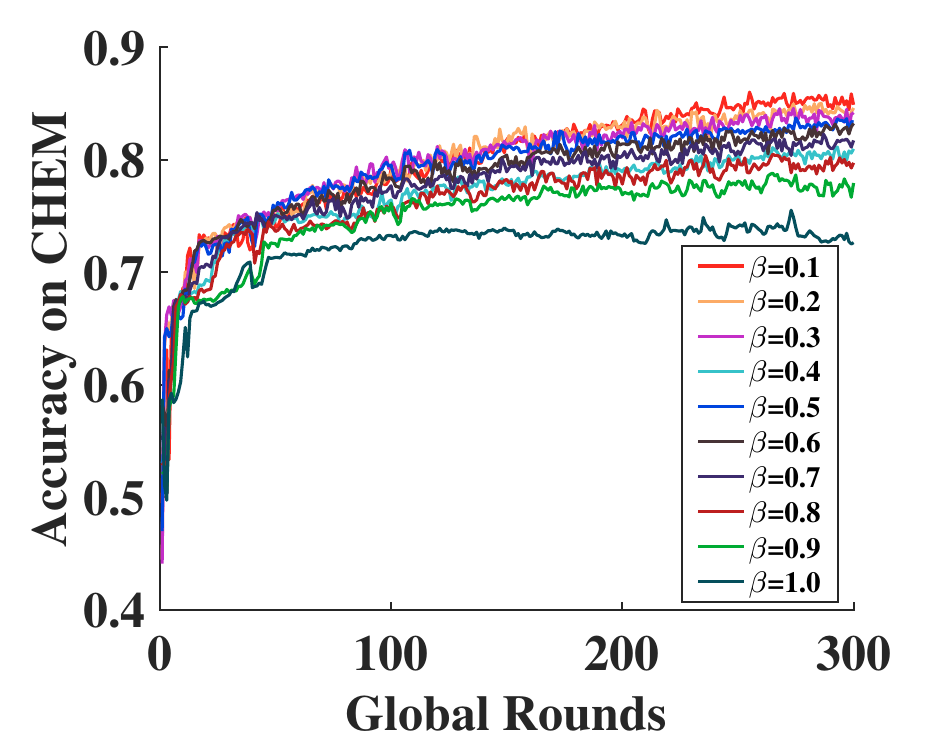}%
\hfill
\includegraphics[width=0.4\textwidth]{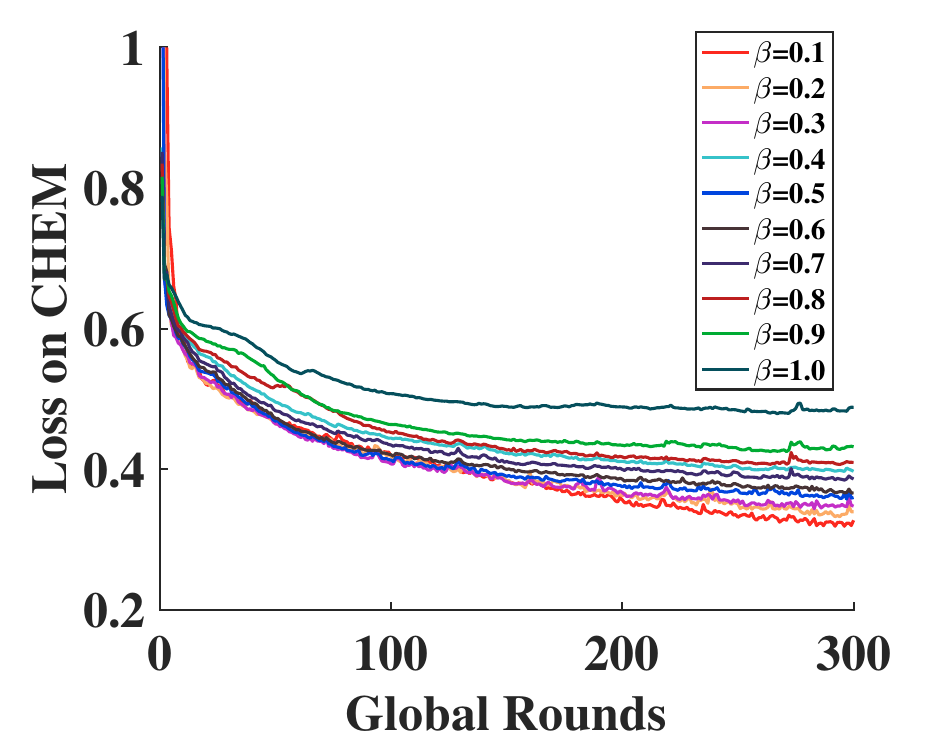}%
\caption{Variation of test accuracy and loss with the number of communication rounds for different sparse ratios $\beta$.}
\label{fig3}
\end{figure}

\subsection{Sparsity Ratios Analysis}
To assess the impact of the sparsity of the personalization component, we fix the low-rank hyperparameter \( \mu \) at 0.001 and vary the sparsity hyperparameter \( \lambda \) using GIN as the backbone. As shown in TABLE \ref{tab3}, a \( \lambda \) value of 0 results in a dense personalization component, leading to unsatisfactory model performance. Dense models tend to overfit the local data, which is particularly problematic in federated learning scenarios with heterogeneous data distributions. As \( \lambda \) increases, the number of parameters in the personalization model decreases, significantly improving accuracy. This improvement can be attributed to sparse regularization, which filters out redundant information, allowing the model to focus on the specific features of the data. We also evaluate sparsity using the Top-$k$ sparsity method, where \( k = \beta \times N_s \) and \( N_s \) denote the total number of parameters in the personalized model. As shown in Figure \ref{fig3}, the optimal performance can be achieved by using only 10\% of the parameters. This suggests that appropriate sparse regularization enhances the performance of models. The model becomes more efficient and generalizable by effectively capturing local knowledge with fewer parameters.

The sparsity in the personalization component ensures that the model enhances the capture of local knowledge, thereby improving model performance. This is crucial in federated learning, where data from different clients can vary significantly. By controlling the sparsity of the personalization component, our method balances model flexibility with the need to maintain generalization across different data sources. In summary, these results highlight the crucial role of sparsity hyperparameters ($\lambda$  or $\beta$) in the efficiency and effectiveness of personalization components in federated learning. This adaptability makes our method an excellent choice for federated learning scenarios with IID and non-IID data distributions.

\begin{table*}[!t]
\renewcommand{\arraystretch}{1.1}
\caption{Ablation experiments for CEFGL. $\bm{W}$, $\bm{S}$ denote low-rank global and sparse personalized components, respectively.}
\label{tab4}
\centering
\resizebox{\textwidth}{!}{
\begin{tabular}{ccccccccccc}
\hline
$\bm{W}$                         & $\bm{S}$                         & CHEM           & SN\_CV         & CHEM\_BIO      & CHEM\_BIO\_SN  & BIO\_SN\_CV    & ALL            & DD             & NCI1           & IMDB-BINARY \\ \hline
- & \checkmark                         & 63.76 & 56.92  & 55.23     & 47.46         & 54.08       & 57.74 & 59.30  & 55.98 & 57.72       \\
\checkmark   &- & 66.42 & 66.38  & 61.16     & 58.62         & 62.15       & 66.03 & 68.54 & 62.46 & 68.81       \\
\checkmark & \checkmark & 83.03 & 74.35  & 74.82     & 74.13         & 72.35       & 75.05 & 74.84 & 68.07 & 75.43       \\ \hline
\end{tabular}}
\end{table*}

\begin{figure*}[!t]
\centering
\includegraphics[width=0.3\textwidth]{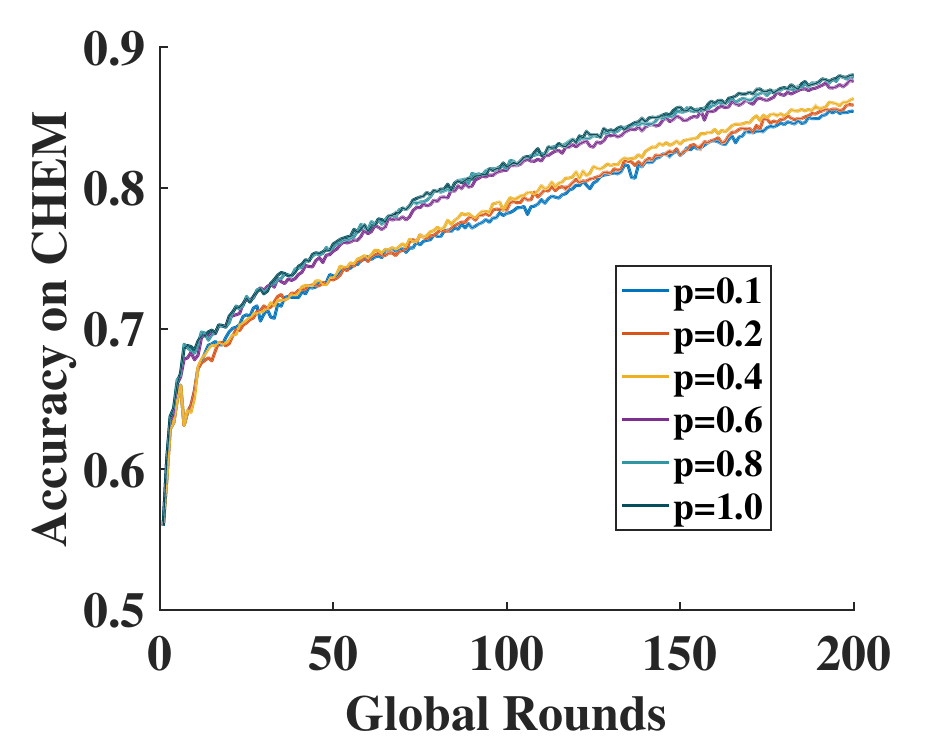}%
\hfill
\includegraphics[width=0.3\textwidth]{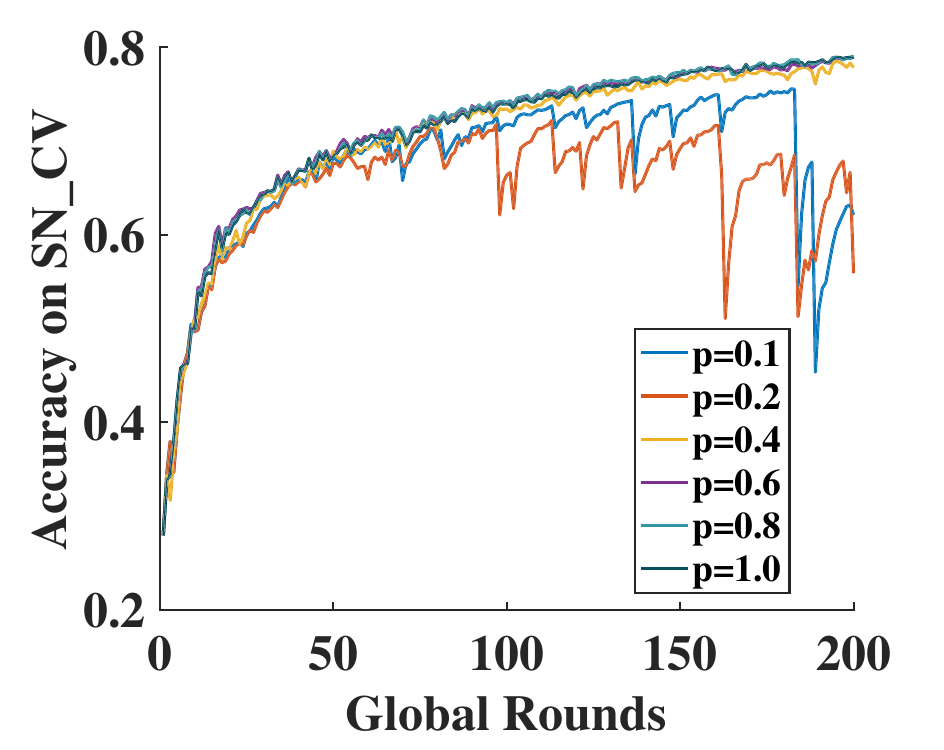}%
\hfill
\includegraphics[width=0.3\textwidth]{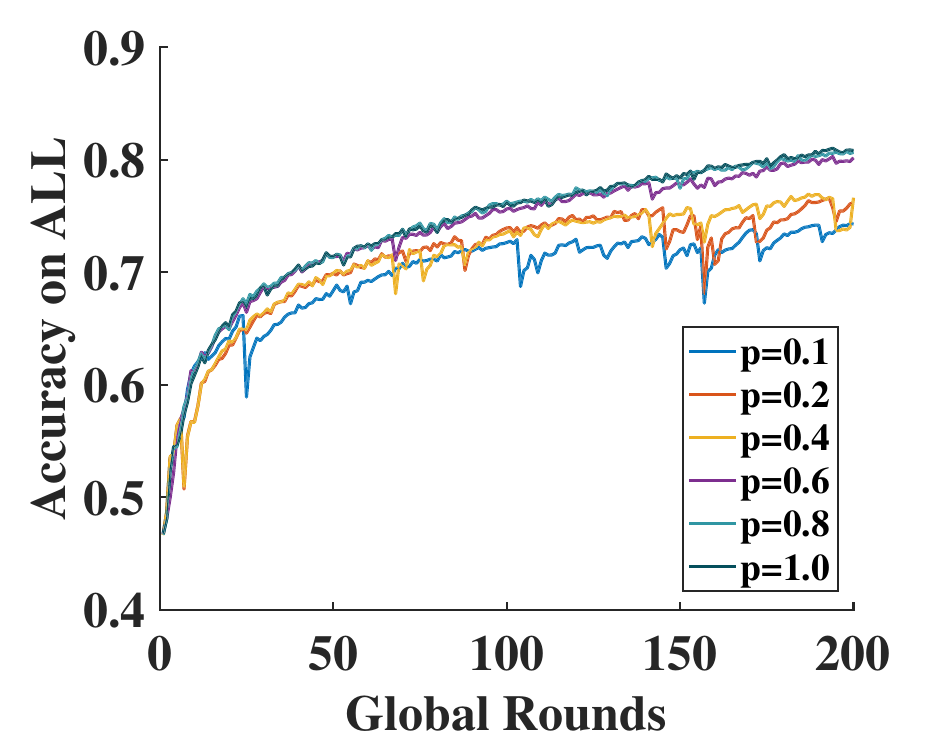}%
\\ 
\includegraphics[width=0.3\textwidth]{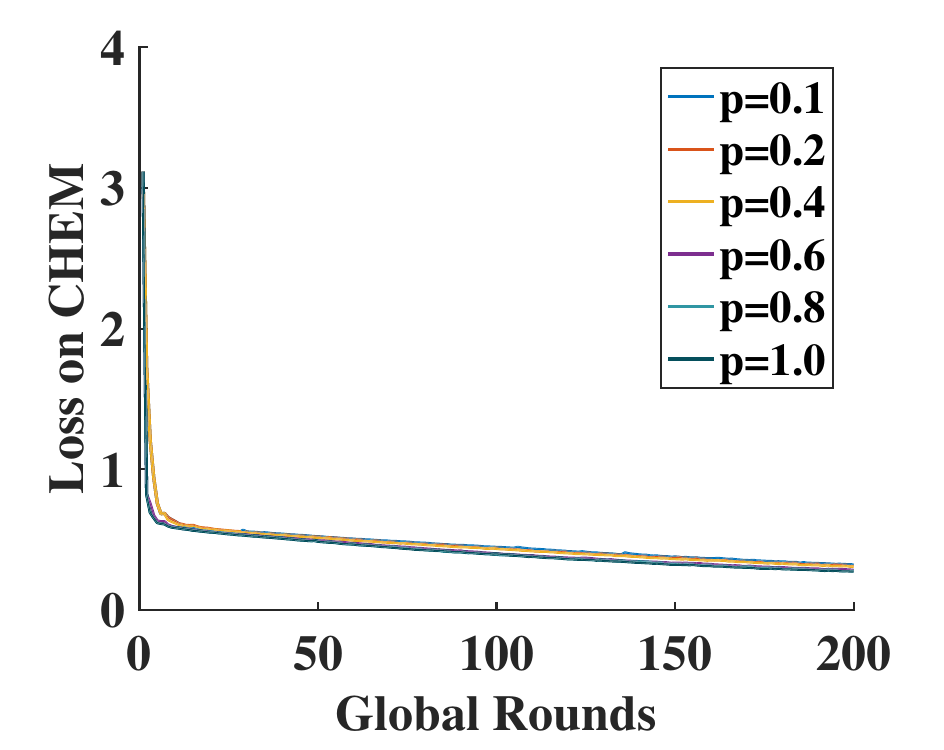}%
\hfill
\includegraphics[width=0.3\textwidth]{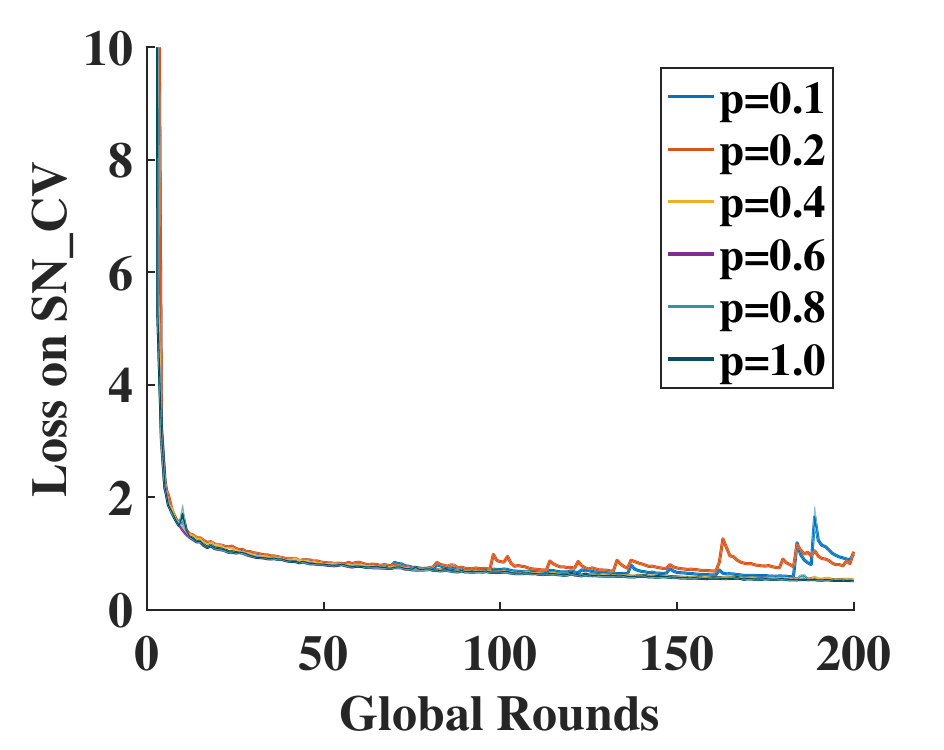}%
\hfill
\includegraphics[width=0.3\textwidth]{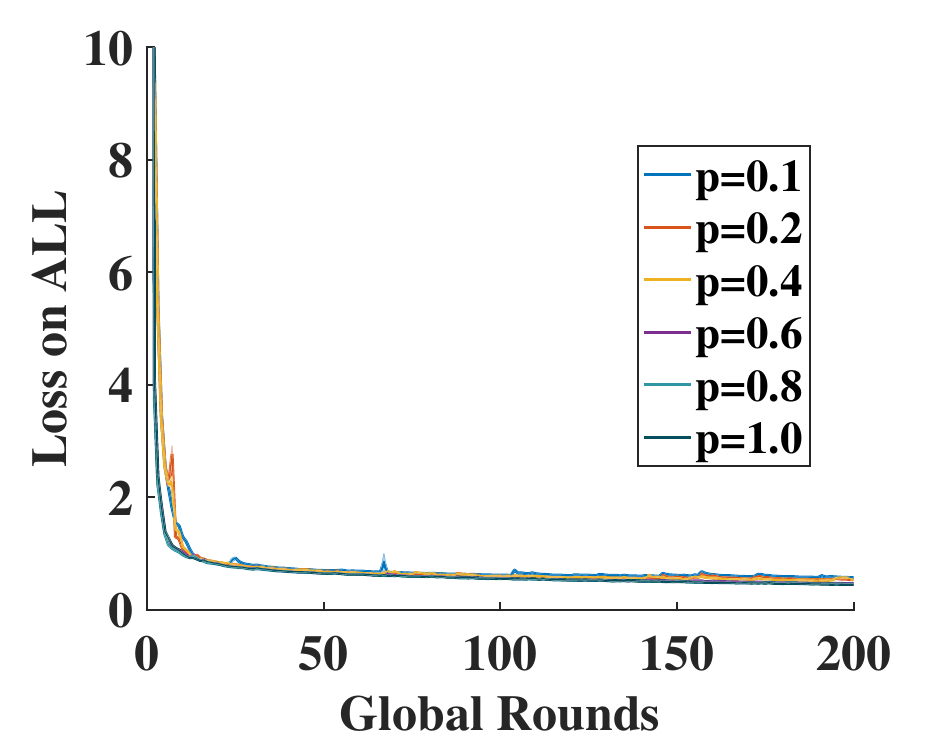}%
\caption{Variation of test accuracy and loss with the number of communication rounds for different communication probabilities $p$.}
\label{fig4}
\end{figure*}

\subsection{Ablation Experiments}
To verify the rationale and validity of the low-rank global component and the sparse personalized component, we compare several variants of CEFGL using GIN as the backbone, with results shown in TABLE \ref{tab4}. When the global component is zeroed out, severe performance is degraded. This is primarily due to the limited number of parameters and the absence of information exchange between clients. In this case, the local model relies solely on the sparse personalization component, which is insufficient for comprehensive learning, especially in a federated setting with diverse data distributions. Conversely, when the personalization components are zeroed out, the model becomes a standard federated graph learning algorithm. This method proves unsuitable for highly heterogeneous cross-dataset scenarios, as it lacks the adaptability provided by personalized components. Although the performance of this pure global component variant shows improvement over the pure personalization component, it still falls short of the hybrid model. The hybrid model, which integrates low-rank global and sparse personalized components, achieves superior performance. This configuration benefits from the complementary strengths of both components. The global component captures common patterns and facilitates information sharing across clients.
In contrast, the personalized component adapts to local data variations, enhancing the overall flexibility and generalization capability of the model. These observations underscore that integrating low-rank global and sparse personalized components is crucial for achieving optimal performance, demonstrating the ability of the model to balance global knowledge extraction and local adaptation. This balanced method is particularly advantageous in federated learning with non-IID data.

\begin{figure}[t]
\centering
\includegraphics[width=0.24\textwidth]{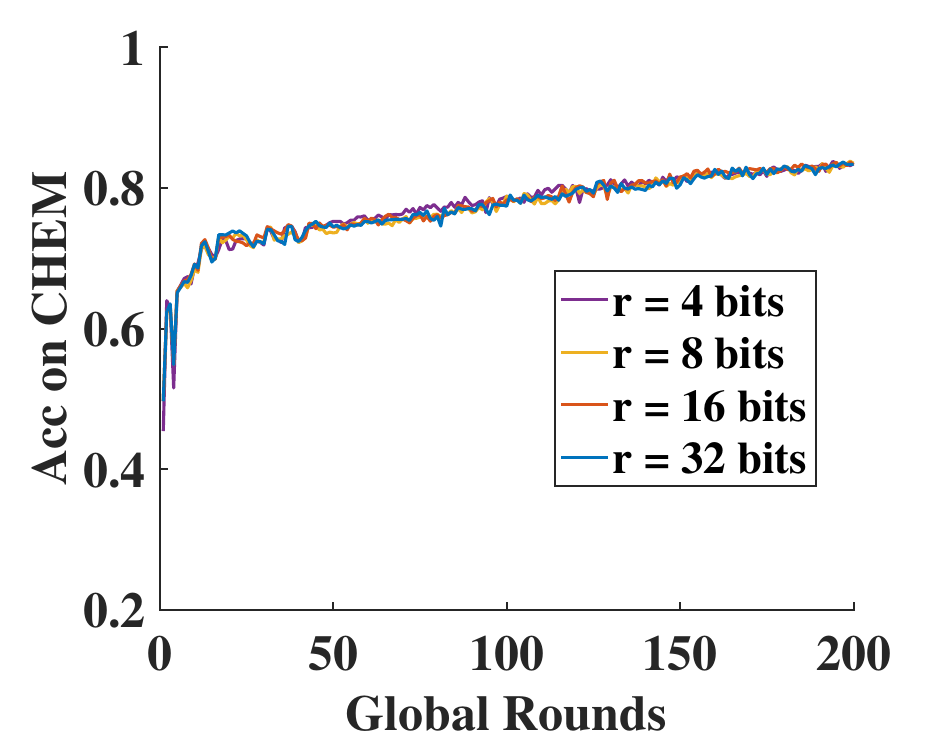}%
\includegraphics[width=0.24\textwidth]{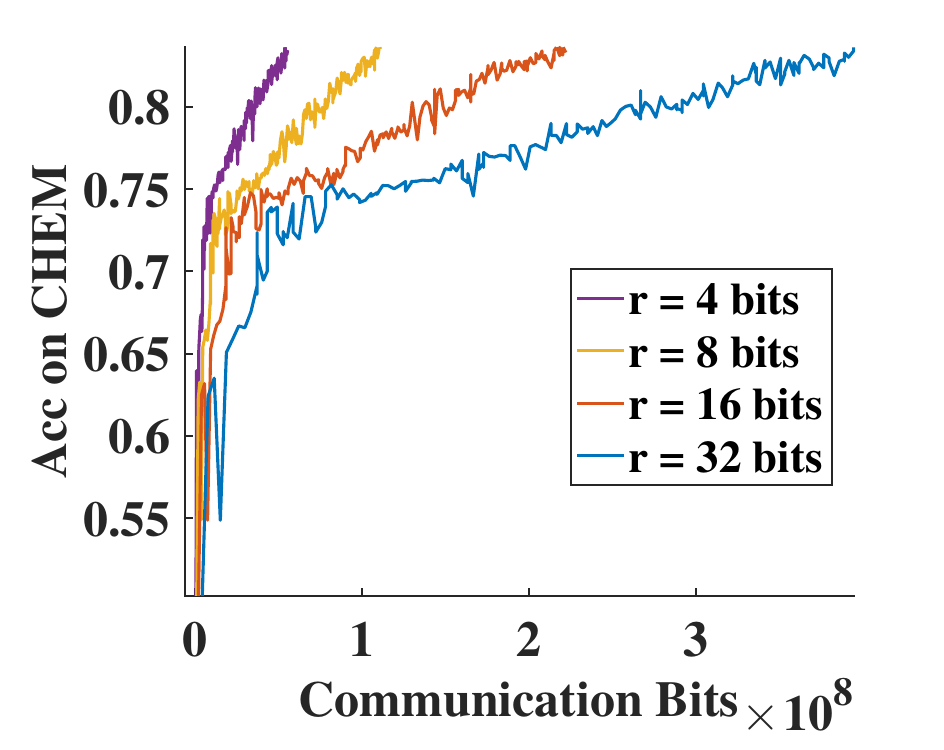}
\includegraphics[width=0.24\textwidth]{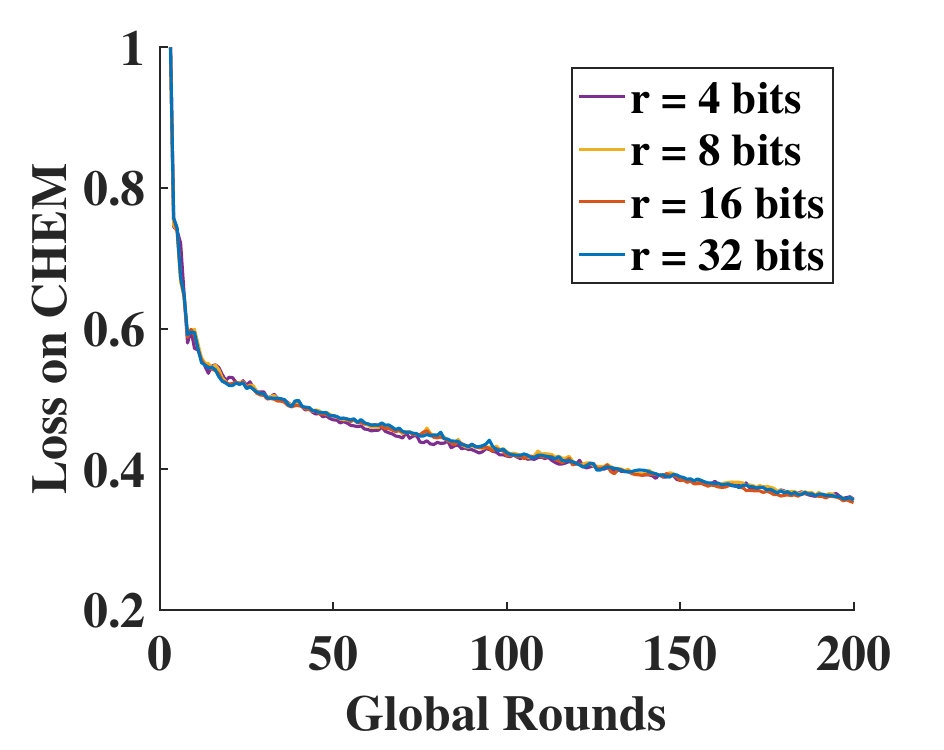}
\includegraphics[width=0.24\textwidth]{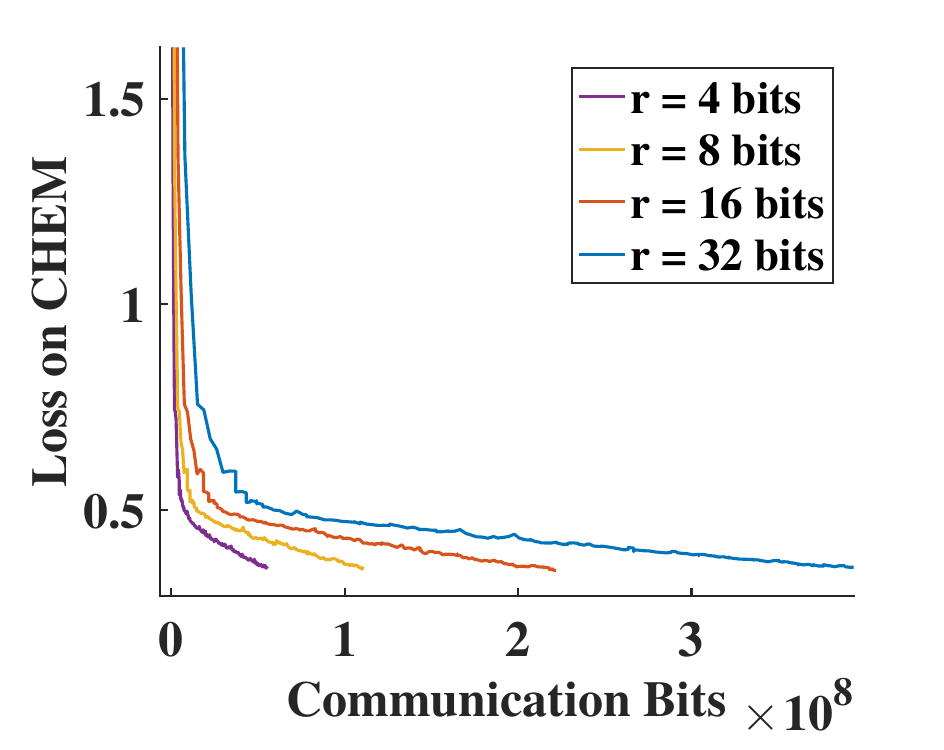}
\caption{Test accuracy and loss of CEFGL after employing quantization compression $Q_r(\cdot)$. The number of quantization bits $r$ is set to $r \in \{4,8,16,32\}$.}
\label{fig5}
\end{figure}

\subsection{Number of Local Iterations}
We explore the effect of varying the expected number of local iterations on model performance. The expected number of localized iterations is defined as \( 1/p \), where \( p \) denotes the communication probability. For instance, \( p = 0.1 \) means that the client communicates with the server, on average, once every ten localized iterations. To understand the impact of this parameter, we investigate the model accuracy and loss for \( p \in \{0.1, 0.2, 0.4, 0.6, 0.8, 1.0\} \). The results, shown in Figure \ref{fig4}, reveal several key insights: 
\begin{itemize}
    \item \textbf{Model Performance and Convergence}: Smaller values of \( p \) (i.e., more localized training) do not significantly affect the overall performance or convergence of the model. The accuracy and loss metrics remain relatively stable across different values of \( p \), indicating that the model is robust to changes in communication frequency. 
    \item \textbf{Communication Cost}: Lowering \( p \) effectively reduces the communication frequency between the client and the server. Reduced communication frequency translates to lower communication costs, which is especially beneficial in federated learning scenarios where communication overhead can be a significant bottleneck.
    \item \textbf{Privacy Considerations}: Variable localized training step sizes prevent the server from accurately recovering the actual gradient of the client, thereby avoiding privacy leakage to some extent.
\end{itemize}

While performance metrics remain stable, it is essential to consider the trade-off between communication frequency and the potential risk of model divergence. Excessive localization (very low \( p \)) may result in the clients' models drifting too far apart before synchronization, potentially affecting long-term convergence. However, this risk appears minimal within the tested range ($0.1 \leq p \leq 1.0$). The experiments demonstrate that lowering the communication probability \( p \) can reduce communication costs and enhance privacy without significantly affecting model performance or convergence. This finding highlights the flexibility and robustness of our method, making it well-suited for federated learning environments where communication efficiency is critical. By adjusting the communication probabilities, we can balance maintaining model accuracy and reducing communication overhead.

\subsection{Quantization}
We investigate the application of quantization compression \( Q_r (\cdot) \) in CEFGL, varying the number of bits as \( r \in \{4, 8, 16, 32\} \). Using the quantization method proposed by Alistarh et al. \cite{alistarh2017qsgd}, we present the results after 200 communication rounds in Figure \ref{fig5}. Using 4-bit quantization, which corresponds to a communication cost of only 12.5\%, does not significantly degrade model performance. Accuracy and convergence speeds remain comparable to those achieved with higher bits quantization levels. It demonstrates that CEFGL effectively retains essential training information even with lower granularity parameters. Additionally, low-bit quantization reduces the amount of data transferred between the client and the server, significantly lowering communication costs.

The proposed method effectively reduces communication overhead through low-bit quantization without sacrificing model performance or convergence speed. This highlights the efficiency of our method, making it well-suited for large-scale federated learning deployments where communication efficiency is critical and enabling implementation in resource-constrained environments such as mobile and edge devices.

\begin{figure}[!t]
\centering
\subfloat[]{\includegraphics[width=0.35\textwidth, height=1.4in]{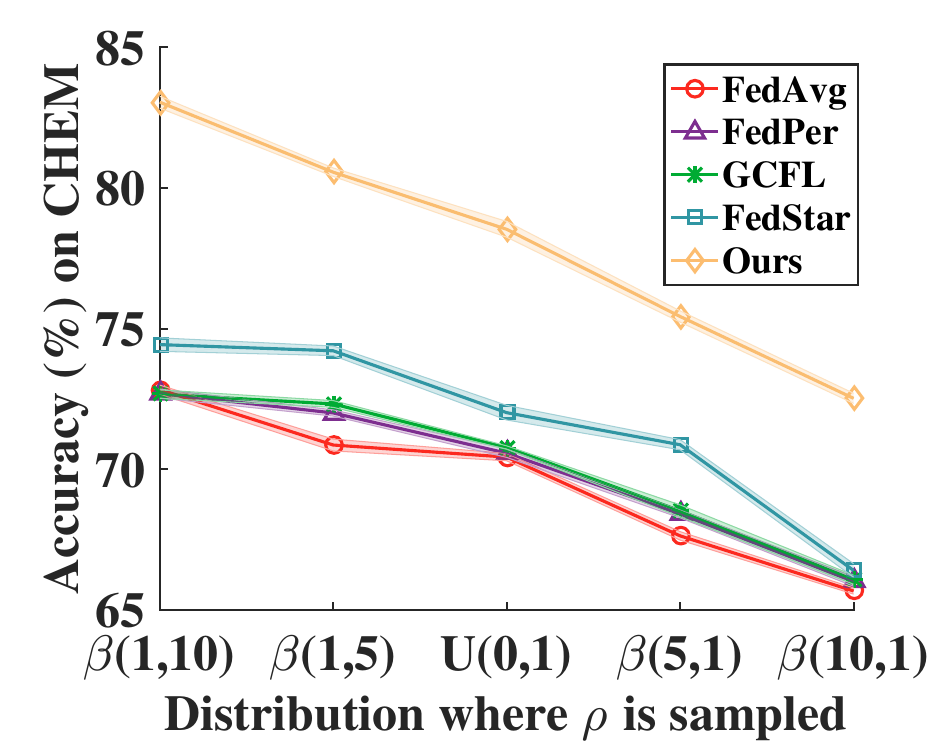}%
\label{fig6_first_case}}
\hfil
\subfloat[]{\includegraphics[width=0.4\textwidth, height=1.6in]{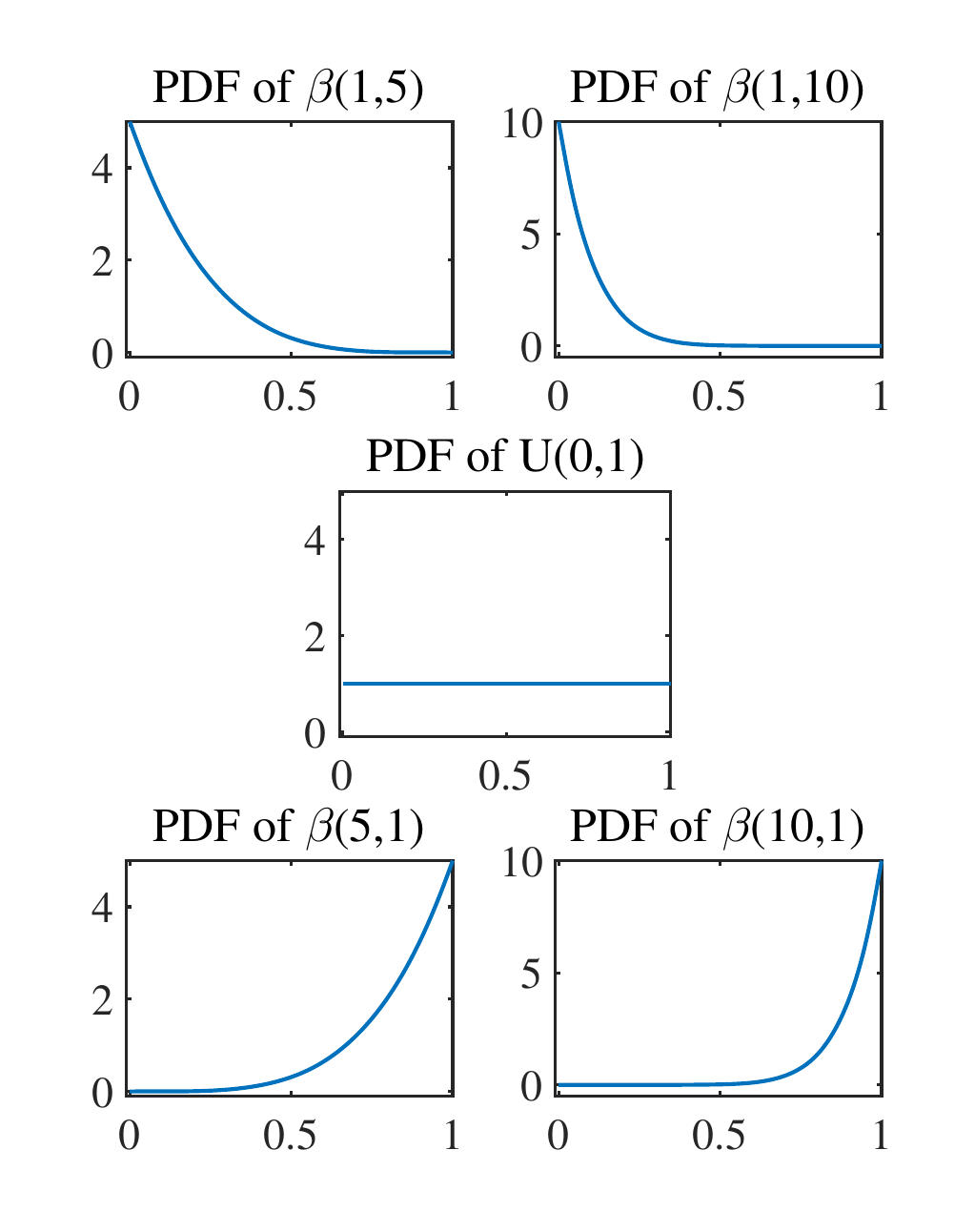}%
\label{fig6_second_case}}
\caption{(a) Accuracy on  CHEM when clients accidentally drop out with different probabilities $\rho$. (b) The distributions' corresponding probability density functions (PDF).}
\label{fig6}
\end{figure}

\subsection{Clients Accidentally Dropping Out}
Due to fluctuating network connections, clients in mobile environments may unexpectedly exit in one iteration and rejoin in another. To evaluate the performance of various FGL methods under these conditions, we conduct experiments as shown in Figure \ref{fig6}. 
Unlike many existing methods with a fixed client drop rate, we sample the drop rate \( \rho \) from a Beta distribution in each iteration. Larger values of \( \rho \) indicate greater instability. The experimental results reveal several significant findings. The accuracy of most FGL methods decreases under unstable conditions. This is expected, as inconsistency in client participation undermines the ability of the model to aggregate information efficiently. However, CEFGL maintains the highest accuracy even under extremely unstable conditions, i.e., when $\rho$ is sampled from $\beta(10,1)$. The ability of CEFGL to maintain its dominance and consistent performance in unstable environments can be attributed to two main factors:
\begin{itemize}
    \item \textbf{Unforced aggregation:} CEFGL does not require aggregation at every iteration, allowing the model to continue learning even when some clients are unavailable.
    \item \textbf{Sparse Personalisation Component}: The sparse personalization component in CEFGL effectively complements global knowledge. By focusing on the critical local information, the model adapts to changing environments without relying solely on continuous client engagement.
\end{itemize}
CEFGL is particularly suitable for FGL environments with unstable clients. It is designed to cope with the challenges posed by network fluctuations and to leverage sparse personalization to enhance the overall robustness and adaptability of the model. This makes CEFGL ideal for deployment in mobile and other dynamic environments where maintaining high accuracy and consistent performance is critical.

\begin{figure*}[!t]
\centering
\subfloat[]{\includegraphics[width=0.3\textwidth]{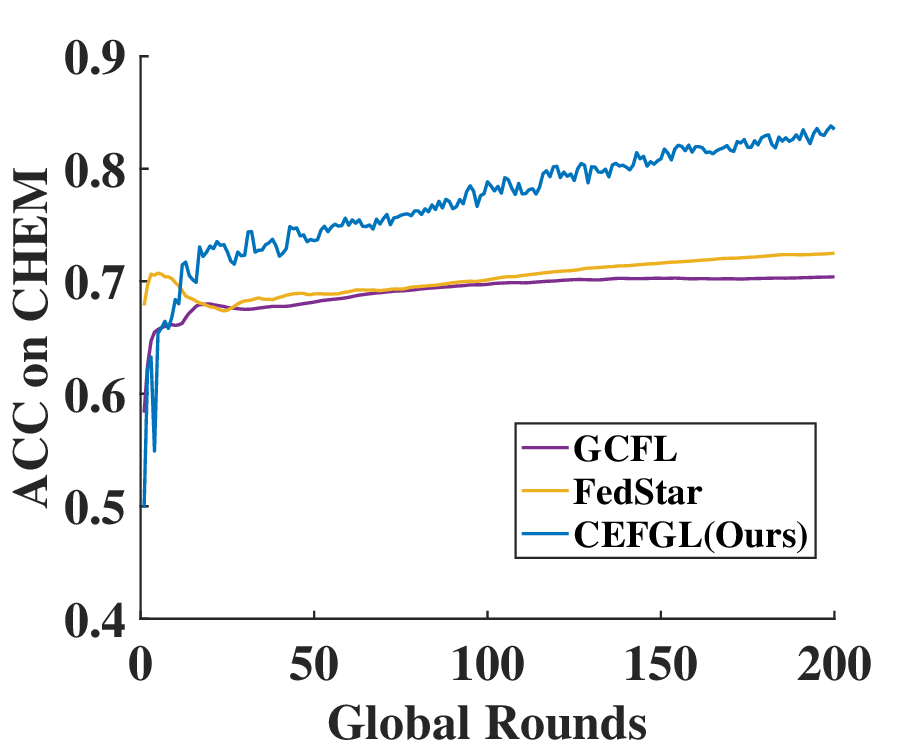}
\label{fig7_1}}%
\hfill
\subfloat[]{\includegraphics[width=0.3\textwidth]{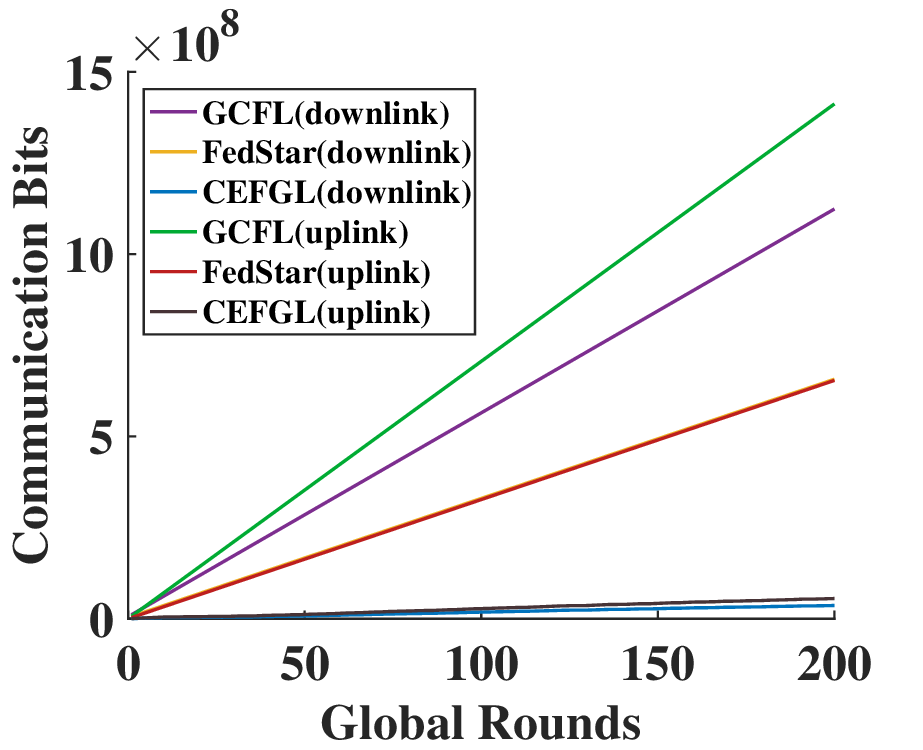}%
\label{fig7_2}}
\hfill
\subfloat[]{\includegraphics[width=0.3\textwidth]{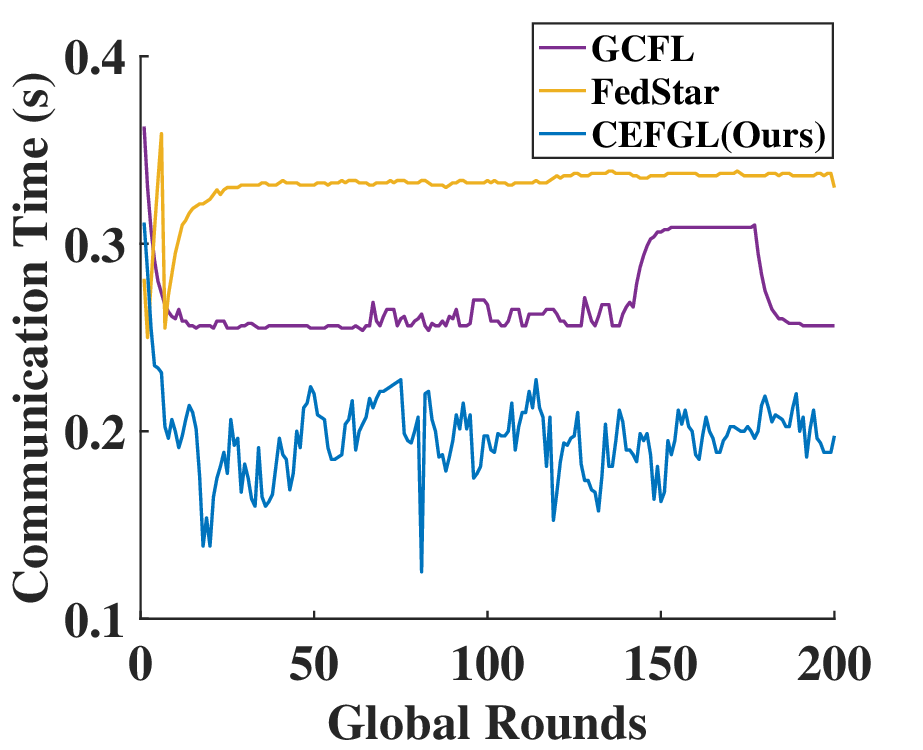}%
\label{fig7_3}}
\caption{(a) Variation of test accuracy and loss with the number of communication rounds. (b) Variation of communication bits with the number of communication rounds. (c) Communication time per global round.}
\label{fig7}
\end{figure*}

\begin{figure}[!t]
\centering
\subfloat[]{\includegraphics[width=0.4\textwidth]{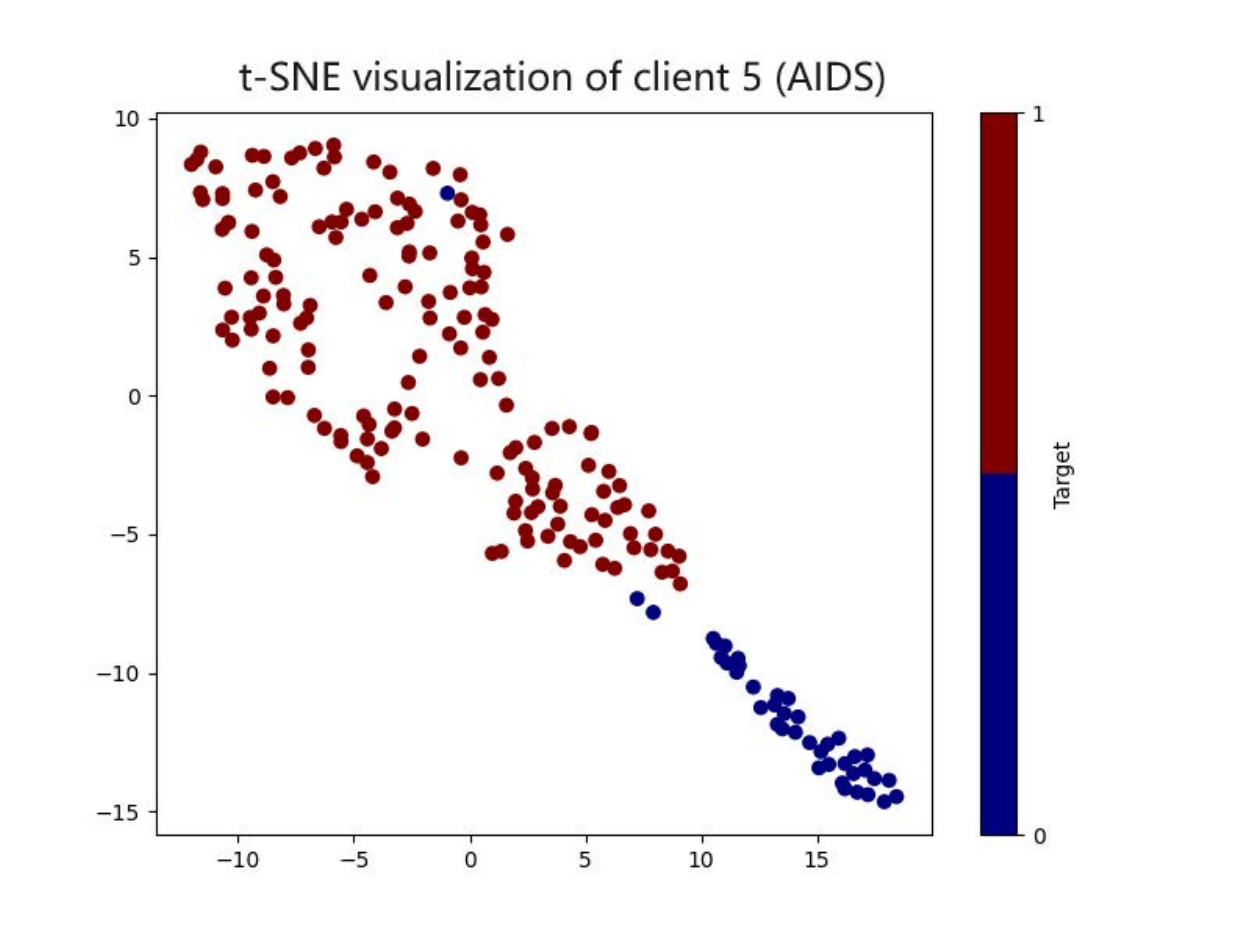}%
\label{fig8_first_case}}
\hfil
\subfloat[]{\includegraphics[width=0.4\textwidth]{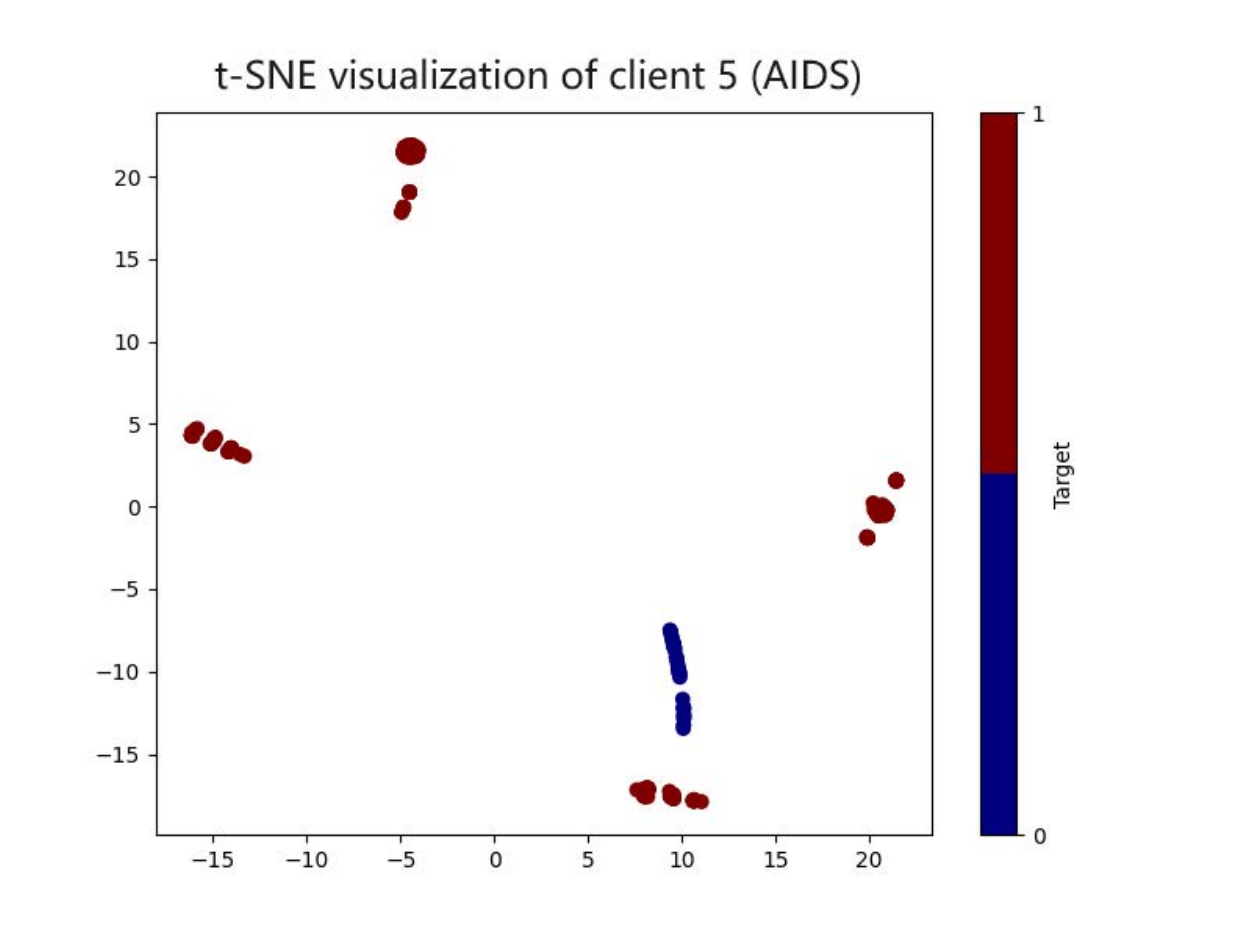}%
\label{fig8_second_case}}
\caption{t-SNE visualization of local graph representations derived from the last hidden layer of the CEFGL trained GIN model. (a) Graph representation from the hybrid model ($\bm{W}$ and $\bm{S}$). (b) Graph representation from the pure global component (only $\bm{W}$).}
\label{fig8}
\end{figure}

\subsection{Further Discussion}
\noindent\textbf{Comparison of Performance and Communications Cost}\quad 
To evaluate the performance and communication cost of the proposed method, we compare various FGL methods under the cross-dataset setting CHEM over 200 rounds of communication. The comparison metrics include model performance (Figure \ref{fig7_1}), the total number of bits transmitted (Figure \ref{fig7_2}), and the wall-clock time per communication round (Figure \ref{fig7_3}). The experimental results demonstrate that the proposed method achieves superior performance with significantly reduced communication costs compared to baseline methods. For example, compared to the optimal method FedStar, SDGRL reduces the number of communication bits by a factor of 18.58, reduces the communication time by a factor of 1.65, and improves the classification accuracy by 5.64\%. In summary, SDGRL consistently delivers the best performance with minimal memory and time consumption. Its ability to reduce communication costs significantly while improving classification accuracy makes it a robust and efficient choice for federated learning in heterogeneous and resource-constrained settings. The superior performance of our method in the cross-dataset setting reaffirms its potential for broader applications in various federated graph learning scenarios.

\noindent\textbf{Visualization}
We use t-SNE \cite{van2008visualizing} to visualize the graph representations extracted from two variants of CEFGL, as shown in Figure \ref{fig8}. The graph representations extracted by the pure global model appear scattered and exhibit significant overlap between samples with different labels. This overlap results in a less distinct classification surface, making it challenging to distinguish between different classes accurately. The scattered distribution indicates that the pure global model struggles to capture the nuanced local variations in the data, leading to suboptimal performance. In contrast, CEFGL incorporating sparse components shows a much more structured representation. Points corresponding to different labels are more distinctly separated. This separation enhances the clarity of the classification boundaries, making it easier for the model to classify the samples accurately. The incorporation of sparse components enables the model to focus on the most critical local information, thereby improving the overall quality of the extracted representations.


\section{Conclusion}
\label{conclusion}
In this work, we propose \textbf{CEFGL}, a communication-efficient personalized federated graph learning algorithm that decomposes model parameters into generic low-rank and private sparse components. We utilize a dual-channel encoder to learn sparse local knowledge in a personalized manner and low-rank global knowledge in a shared manner. To further optimize communication efficiency, we perform multiple local stochastic gradient descent iterations between communication stages and integrate efficient compression techniques into the algorithm. Extensive experiments demonstrate the superiority and communication efficiency of our method in highly heterogeneous cross-dataset scenarios. In future work, we plan to assess the extent of data leakage in the proposed method and enhance its security features by adopting advanced privacy-preserving techniques to protect sensitive information.

\section*{Acknowledgments}
This work is supported in part by the National Natural Science Foundation of China (No.62106259, No.62076234), Beijing Outstanding Young Scientist Program (NO.BJJWZYJH012019100020098), and Beijing Natural Science Foundation (No. 4222029).


\bibliographystyle{elsarticle-num} 
\bibliography{ref}

\begin{thebibliography}{10}
\expandafter\ifx\csname url\endcsname\relax
  \def\url#1{\texttt{#1}}\fi
\expandafter\ifx\csname urlprefix\endcsname\relax\def\urlprefix{URL }\fi
\expandafter\ifx\csname href\endcsname\relax
  \def\href#1#2{#2} \def\path#1{#1}\fi

\bibitem{kipf2016semi}
T.~N. Kipf, M.~Welling, Semi-supervised classification with graph convolutional networks, CoRR abs/1609.02907 (2016).

\bibitem{liu2024aswt}
R.~Liu, R.~Yin, Y.~Liu, W.~Wang, Aswt-sgnn: Adaptive spectral wavelet transform-based self-supervised graph neural network, in: Proceedings of the AAAI Conference on Artificial Intelligence, Vol.~38, 2024, pp. 13990--13998.

\bibitem{Su2023}
X.~Su, Z.~You, D.~Huang, L.~Wang, L.~Wong, B.~Ji, B.~Zhao, Biomedical knowledge graph embedding with capsule network for multi-label drug-drug interaction prediction, IEEE Transactions on Knowledge and Data Engineering 35~(6) (2023) 5640--5651.

\bibitem{LIU2024102563}
J.~Liu, F.~Hu, Q.~Zou, P.~Tiwari, H.~Wu, Y.~Ding, Drug repositioning by multi-aspect heterogeneous graph contrastive learning and positive-fusion negative sampling strategy, Information Fusion 112 (2024) 102563.

\bibitem{WU2023101909}
Y.~Wu, Y.~Chen, Z.~Yin, W.~Ding, I.~King, A survey on graph embedding techniques for biomedical data: Methods and applications, Information Fusion 100 (2023) 101909.

\bibitem{ding2023lggnet}
Y.~Ding, N.~Robinson, C.~Tong, Q.~Zeng, C.~Guan, Lggnet: Learning from local-global-graph representations for brain–computer interface, IEEE Transactions on Neural Networks and Learning Systems 35~(7) (2024) 9773--9786.

\bibitem{GORRIZ2023101945}
J.~Górriz, I.~Álvarez Illán, A.~Álvarez Marquina, J.~Arco, M.~Atzmueller, F.~Ballarini, E.~Barakova, G.~Bologna, P.~Bonomini, G.~Castellanos-Dominguez, D.~Castillo-Barnes, S.~Cho, R.~Contreras, J.~Cuadra, E.~Domínguez, F.~Domínguez-Mateos, R.~Duro, D.~Elizondo, A.~Fernández-Caballero, E.~Fernandez-Jover, M.~Formoso, N.~Gallego-Molina, J.~Gamazo, J.~G. González, J.~Garcia-Rodriguez, C.~Garre, J.~Garrigós, A.~Gómez-Rodellar, P.~Gómez-Vilda, M.~Graña, B.~Guerrero-Rodriguez, S.~Hendrikse, C.~Jimenez-Mesa, M.~Jodra-Chuan, V.~Julian, G.~Kotz, K.~Kutt, M.~Leming, J.~{de Lope}, B.~Macas, V.~Marrero-Aguiar, J.~Martinez, F.~Martinez-Murcia, R.~Martínez-Tomás, J.~Mekyska, G.~Nalepa, P.~Novais, D.~Orellana, A.~Ortiz, D.~Palacios-Alonso, J.~Palma, A.~Pereira, P.~Pinacho-Davidson, M.~Pinninghoff, M.~Ponticorvo, A.~Psarrou, J.~Ramírez, M.~Rincón, V.~Rodellar-Biarge, I.~Rodríguez-Rodríguez, P.~Roelofsma, J.~Santos, D.~Salas-Gonzalez, P.~Salcedo-Lagos, F.~Segovia, A.~Shoeibi, M.~Silva, D.~Simic, J.~Suckling,
  J.~Treur, A.~Tsanas, R.~Varela, S.~Wang, W.~Wang, Y.~Zhang, H.~Zhu, Z.~Zhu, J.~Ferrández-Vicente, Computational approaches to explainable artificial intelligence: Advances in theory, applications and trends, Information Fusion 100 (2023) 101945.

\bibitem{liu2024unbiased}
R.~Liu, R.~Yin, Y.~Liu, W.~Wang, Unbiased and augmentation-free self-supervised graph representation learning, Pattern Recognition 149 (2024) 110274.

\bibitem{LI2024102837}
Y.~Li, J.~Huan, J.~Shen, L.~Chen, J.~Cao, Y.~Cheng, Social network large-scale group decision-making considering dynamic trust relationships and historical preferences of decision makers in opinion evolution, Information Fusion (2024) 102837.

\bibitem{PENG2025102729}
H.~Peng, W.~Zeng, J.~Tang, M.~Wang, H.~Huang, X.~Zhao, Open knowledge graph completion with negative-aware representation learning and multi-source reliability inference, Information Fusion 115 (2025) 102729.

\bibitem{ZHANG2024102581}
Z.~Zhang, L.~Bai, L.~Zhu, Ts-align: A temporal similarity-aware entity alignment model for temporal knowledge graphs, Information Fusion 112 (2024) 102581.

\bibitem{10114977}
W.~Fan, X.~Zhao, Q.~Li, T.~Derr, Y.~Ma, H.~Liu, J.~Wang, J.~Tang, Adversarial attacks for black-box recommender systems via copying transferable cross-domain user profiles, IEEE Transactions on Knowledge and Data Engineering 35~(12) (2023) 12415--12429.

\bibitem{PAZRUZA2024102497}
J.~Paz-Ruza, A.~Alonso-Betanzos, B.~Guijarro-Berdiñas, B.~Cancela, C.~Eiras-Franco, Sustainable transparency on recommender systems: Bayesian ranking of images for explainability, Information Fusion 111 (2024) 102497.

\bibitem{HIMEUR20211}
Y.~Himeur, A.~Alsalemi, A.~Al-Kababji, F.~Bensaali, A.~Amira, C.~Sardianos, G.~Dimitrakopoulos, I.~Varlamis, A survey of recommender systems for energy efficiency in buildings: Principles, challenges and prospects, Information Fusion 72 (2021) 1--21.

\bibitem{ZHU2025102703}
W.~Zhu, X.~Zhou, S.~Lan, W.~Wang, Z.~Hou, Y.~Ren, T.~Pan, A dual branch graph neural network based spatial interpolation method for traffic data inference in unobserved locations, Information Fusion 114 (2025) 102703.

\bibitem{XU2024102292}
D.~Xu, H.~Peng, Y.~Tang, H.~Guo, Hierarchical spatio-temporal graph convolutional neural networks for traffic data imputation, Information Fusion 106 (2024) 102292.

\bibitem{de2023guide}
J.~W. de~Kok, M.~{\'A}.~A. de~la Hoz, Y.~de~Jong, V.~Brokke, P.~W. Elbers, P.~Thoral, A.~Castillejo, T.~Trenor, J.~M. Castellano, A.~E. Bronchalo, et~al., A guide to sharing open healthcare data under the general data protection regulation, Scientific data 10~(1) (2023) 404.

\bibitem{wu2021towards}
K.~Wu, Towards a universal cognitive tool: designing accessible visualization for people with intellectual and developmental disabilities, ACM SIGACCESS Accessibility and Computing~(131) (2021) 1--6.

\bibitem{mcmahan2017communication}
B.~McMahan, E.~Moore, D.~Ramage, S.~Hampson, B.~A. y~Arcas, Communication-efficient learning of deep networks from decentralized data, in: Artificial intelligence and statistics, PMLR, 2017, pp. 1273--1282.

\bibitem{SABAH2024102834}
F.~Sabah, Y.~Chen, Z.~Yang, A.~Raheem, M.~Azam, N.~Ahmad, R.~Sarwar, Communication optimization techniques in personalized federated learning: Applications, challenges and future directions, Information Fusion (2024) 102834.

\bibitem{HUANG2024102576}
W.~Huang, D.~Wang, X.~Ouyang, J.~Wan, J.~Liu, T.~Li, Multimodal federated learning: Concept, methods, applications and future directions, Information Fusion 112 (2024) 102576.

\bibitem{10056291}
C.~Wang, B.~Chen, G.~Li, H.~Wang, Automated graph neural network search under federated learning framework, IEEE Transactions on Knowledge and Data Engineering 35~(10) (2023) 9959--9972.

\bibitem{HU2024102042}
H.~xuan Hu, C.~Cao, Q.~Hu, Y.~Zhang, Federated learning enabled graph convolutional autoencoder and factorization machine for potential friendship prediction in social networks, Information Fusion 102 (2024) 102042.

\bibitem{xie2021federated}
H.~Xie, J.~Ma, L.~Xiong, C.~Yang, Federated graph classification over non-iid graphs, Advances in neural information processing systems 34 (2021) 18839--18852.

\bibitem{10241965}
Z.~Ye, X.~Zhang, X.~Chen, H.~Xiong, D.~Yu, Adaptive clustering based personalized federated learning framework for next poi recommendation with location noise, IEEE Transactions on Knowledge and Data Engineering 36~(5) (2024) 1843--1856.

\bibitem{karimireddy2019scaffold}
S.~P. Karimireddy, S.~Kale, M.~Mohri, S.~J. Reddi, S.~U. Stich, A.~T. Suresh, Scaffold: Stochastic controlled averaging for on-device federated learning, CoRR abs/1910.06378 (2019).

\bibitem{collins2021exploiting}
L.~Collins, H.~Hassani, A.~Mokhtari, S.~Shakkottai, Exploiting shared representations for personalized federated learning, in: International conference on machine learning, PMLR, 2021, pp. 2089--2099.

\bibitem{t2020personalized}
C.~T~Dinh, N.~Tran, J.~Nguyen, Personalized federated learning with moreau envelopes, Advances in Neural Information Processing Systems 33 (2020) 21394--21405.

\bibitem{li2020federated}
T.~Li, A.~K. Sahu, M.~Zaheer, M.~Sanjabi, A.~Talwalkar, V.~Smith, Federated optimization in heterogeneous networks, Proceedings of Machine learning and systems 2 (2020) 429--450.

\bibitem{luo2022adapt}
J.~Luo, S.~Wu, Adapt to adaptation: Learning personalization for cross-silo federated learning, in: The 31st International Joint Conference on Artificial Intelligence., Vol. 2022, 2022, p. 2166.

\bibitem{huang2023fusion}
T.~Huang, L.~Shen, Y.~Sun, W.~Lin, D.~Tao, Fusion of global and local knowledge for personalized federated learning, CoRR abs/2302.11051 (2023).

\bibitem{tan2022towards}
A.~Z. Tan, H.~Yu, L.~Cui, Q.~Yang, Towards personalized federated learning, IEEE Transactions on Neural Networks and Learning Systems 34~(12) (2023) 9587--9603.

\bibitem{mishchenko2022proxskip}
K.~Mishchenko, G.~Malinovsky, S.~Stich, P.~Richt{\'a}rik, Proxskip: Yes! local gradient steps provably lead to communication acceleration! finally!, in: International Conference on Machine Learning, PMLR, 2022, pp. 15750--15769.

\bibitem{10423871}
X.~Yang, H.~Yu, X.~Gao, H.~Wang, J.~Zhang, T.~Li, Federated continual learning via knowledge fusion: A survey, IEEE Transactions on Knowledge and Data Engineering (2024) 1--20.

\bibitem{huang2021personalized}
Y.~Huang, L.~Chu, Z.~Zhou, L.~Wang, J.~Liu, J.~Pei, Y.~Zhang, Personalized cross-silo federated learning on non-iid data, in: Proceedings of the AAAI conference on artificial intelligence, Vol.~35, 2021, pp. 7865--7873.

\bibitem{he2022spreadgnn}
C.~He, E.~Ceyani, K.~Balasubramanian, M.~Annavaram, S.~Avestimehr, Spreadgnn: Decentralized multi-task federated learning for graph neural networks on molecular data, in: Proceedings of the AAAI Conference on Artificial Intelligence, Vol.~36, 2022, pp. 6865--6873.

\bibitem{9606516}
F.~Chen, P.~Li, T.~Miyazaki, C.~Wu, Fedgraph: Federated graph learning with intelligent sampling, IEEE Transactions on Parallel and Distributed Systems 33~(8) (2022) 1775--1786.

\bibitem{ghosh2020efficient}
A.~Ghosh, J.~Chung, D.~Yin, K.~Ramchandran, An efficient framework for clustered federated learning, Advances in Neural Information Processing Systems 33 (2020) 19586--19597.

\bibitem{zhang2022graph}
C.~Zhang, S.~Zhang, S.~Yu, J.~James, Graph-based traffic forecasting via communication-efficient federated learning, in: 2022 IEEE Wireless Communications and Networking Conference (WCNC), IEEE, 2022, pp. 2041--2046.

\bibitem{tan2023federated}
Y.~Tan, Y.~Liu, G.~Long, J.~Jiang, Q.~Lu, C.~Zhang, Federated learning on non-iid graphs via structural knowledge sharing, in: Proceedings of the AAAI conference on artificial intelligence, Vol.~37, 2023, pp. 9953--9961.

\bibitem{yao2024fedgcn}
Y.~Yao, W.~Jin, S.~Ravi, C.~Joe-Wong, Fedgcn: Convergence-communication tradeoffs in federated training of graph convolutional networks, Advances in neural information processing systems 36 (2024).

\bibitem{candes2011robust}
E.~J. Cand{\`e}s, X.~Li, Y.~Ma, J.~Wright, Robust principal component analysis?, Journal of the ACM (JACM) 58~(3) (2011) 1--37.

\bibitem{arivazhagan2019federated}
M.~G. Arivazhagan, V.~Aggarwal, A.~K. Singh, S.~Choudhary, Federated learning with personalization layers, CoRR abs/1912.00818 (2019).

\bibitem{XuHLJ19}
K.~Xu, W.~Hu, J.~Leskovec, S.~Jegelka, How powerful are graph neural networks?, in: 7th International Conference on Learning Representations, {ICLR} 2019, New Orleans, LA, USA, May 6-9, 2019, OpenReview.net, 2019.

\bibitem{HamiltonYL17}
W.~L. Hamilton, Z.~Ying, J.~Leskovec, Inductive representation learning on large graphs, in: I.~Guyon, U.~von Luxburg, S.~Bengio, H.~M. Wallach, R.~Fergus, S.~V.~N. Vishwanathan, R.~Garnett (Eds.), Advances in Neural Information Processing Systems 30: Annual Conference on Neural Information Processing Systems 2017, December 4-9, 2017, Long Beach, CA, {USA}, 2017, pp. 1024--1034.

\bibitem{alistarh2017qsgd}
D.~Alistarh, D.~Grubic, J.~Li, R.~Tomioka, M.~Vojnovic, Qsgd: Communication-efficient sgd via gradient quantization and encoding, Advances in neural information processing systems 30 (2017).

\bibitem{van2008visualizing}
L.~Van~der Maaten, G.~Hinton, Visualizing data using t-sne., Journal of machine learning research 9~(11) (2008).

\end{thebibliography}

\end{document}